\RequirePackage{fix-cm}
\documentclass[twocolumn]{svjour3}          
\smartqed  
\usepackage{booktabs}
\usepackage{amsmath}
\usepackage{amssymb}
\usepackage{mathtools}
\usepackage{xspace}
\usepackage{hyperref}
\usepackage{graphicx}
\usepackage{caption}
\usepackage{subcaption}
\usepackage{multirow}
\usepackage{color, colortbl}
\usepackage{fancyvrb}
\usepackage{xcolor}
\usepackage{appendix}
\usepackage{natbib}
\usepackage{comment}
\usepackage{pifont}
\usepackage{bbm}

\newcommand{\eg}{\emph{e.g.,}\xspace}

\newcommand{\ie}{\emph{i.e.,}\xspace}

\DeclareMathOperator*{\argmin}{arg\,min}
\DeclareMathOperator*{\arcosh}{arcosh}
\DeclareMathOperator*{\artanh}{artanh}

\newcommand{\newpaper}[1]{\textcolor{black}{#1}}
\newcommand{\newpapertwo}[1]{\textcolor{black}{#1}}
\newcommand{\lastpaper}[1]{\textcolor{black}{#1}}

\definecolor{Gray}{gray}{0.9}
\definecolor{pink}{rgb}{1.0, 0.85, 0.85}

\journalname{International Journal of Computer Vision}

\hypersetup{pdfstartview={FitH}}
\hypersetup{pdfborder={0 0 0}}
\hypersetup{pdfpagemode={UseNone}}
\hypersetup{colorlinks=true}
\hypersetup{citecolor=blue}
\hypersetup{linkcolor=blue}

\begin{document}

\title{Hyperbolic Deep Learning in Computer Vision: A Survey}

\author{Pascal Mettes \and Mina Ghadimi Atigh \and Martin Keller-Ressel \and\\Jeffrey Gu \and Serena Yeung
}

\institute{Pascal Mettes \at
University of Amsterdam, the Netherlands\\
\email{p.s.m.mettes@uva.nl}
\and
Mina Ghadimi Atigh \at
University of Amsterdam, the Netherlands\\
\email{m.ghadimiatigh@uva.nl}
\and
Martin Keller-Ressel \at
TU Dresden, Germany\\
\email{martin.keller-ressel@tu-dresden.de}
\and
Jeffrey Gu \at
Stanford University, USA\\
\email{jeffgu@stanford.edu}
\and
Serena Yeung \at
Stanford University, USA\\
\email{syyeung@stanford.edu}
}

\date{Received: date / Accepted: date}

\maketitle

\begin{abstract}
Deep representation learning is a ubiquitous part of modern computer vision. While Euclidean space has been the de facto standard manifold for learning visual representations, hyperbolic space has recently gained rapid traction for learning in computer vision. Specifically, hyperbolic learning has shown a strong potential to embed hierarchical structures, learn from limited samples, quantify uncertainty, add robustness, limit error severity, and more. In this paper, we provide a categorization and in-depth overview of current literature on hyperbolic learning for computer vision. We research both supervised and unsupervised literature and identify three main research themes in each direction. We outline how hyperbolic learning is performed in all themes and discuss the main research problems that benefit from current advances in hyperbolic learning for computer vision. Moreover, we provide a high-level intuition behind hyperbolic geometry and outline open research questions to further advance research in this direction.
\end{abstract}
\section{Introduction}
From image segmentation to future frame prediction and from video grounding to generating images, deep representation learning is the central component that drives modern computer vision problems \citep{lecun2015deep}. In short succession, many differentiable layers and network architectures have been proposed to tackle visual research problems \citep{gu2018recent,bommasani2021opportunities,khan2022transformers}. While different in structure, scope, and inductive biases, all are based on Euclidean operators and therefore - implicitly or explicitly - assume that data is best represented on regular grids.

Euclidean space forms an intuitive and grounded underlying manifold, but its inherent properties are not a best match for all types of data. Consider for example hierarchical structures such as trees, ontologies, and taxonomies. Hierarchies are foundational building blocks across all scientific disciplines to formalize our knowledge \citep{noy1997state}. In hierarchies, the number of nodes grows exponentially with depth, from few coarse-grained to many fine-grained nodes. The volume of a ball in Euclidean space however, grows only polynomially with its diameter. An alternative geometry is needed to match the nature of hierarchies.

In the quest for a more appropriate geometry of hierarchies, hyperbolic geometry provides a direct fit \citep{bridson2013metric}.
In essence, hyperbolic and Euclidean geometry are different in only one aspect: the parallel postulate. In Euclidean space, there is exactly one parallel line that goes through a point not on the other line. In hyperbolic space, there are at least two such parallel lines. This change comes with many consequences and as a result, hyperbolic geometry can be seen
as a geometry of constant negative curvature. In the context of deep learning this geometry has many attractive properties, such as its hierarchical structure and exponential expansion.

Empowered by these geometric properties, hierarchical embeddings have in recent years been performed in hyperbolic space with great success \citep{nickel2017poincare}, leading to unparalleled abilities to embed deep and complex trees with minimal distortion \citep{ganea2018hyperbolic2,sala2018representation}. This has led to rapid advances in hyperbolic deep learning across many disciplines and research areas, including but not limited to graph networks \citep{chami2019hyperbolic,liu2019hyperbolic,dai2021hyperbolic}, text embeddings~\citep{tifrea2019poincar,zhu2020hypertext}, molecular representation learning \citep{klimovskaia2020poincare,yu2020semi,wu2021hyperbolic}, and recommender systems \citep{mirvakhabova2020performance,wang2021fully,yang2022hrcf}.

In the wake of other disciplines, computer vision has in recent years also benefited from research into deep learning in hyperbolic space. A quickly growing body of literature has shown that hyperbolic embeddings benefit few-shot learning, zero-shot recognition, out-of-distribution generalization, uncertainty quantification, generative learning, and hierarchical representation learning amongst others. These works show evidence that hyperbolic geometry has a lot of potential for learning in computer vision.

This survey provides an in-depth overview and categorization of the recent boom in hyperbolic computer vision literature. These works have investigated hyperbolic learning across many visual research problems with different solutions. As a result, it is unclear how current literature is connected, what is common and new in each work, and in which direction the field is heading. This survey seeks to fill this void.  We investigate both supervised and unsupervised papers. For supervised learning, we identify three shared themes amongst current papers, where samples are matched to either gyroplanes, prototypes, or other samples in hyperbolic space. For unsupervised papers, we dive into the three main axes explored in current papers, namely generative learning, clustering, and self-supervised learning. \cite{peng2021hyperbolic} have recently written a general survey on hyperbolic neural networks but their scope did not include the computer vision literature on hyperbolic learning. This survey fills this void.

The rest of the paper is organised as follows. In Section \ref{sec:background} we provide the background on hyperbolic geometry and foundational papers on hyperbolic embeddings and hyperbolic neural networks. Sections \ref{sec:supervised} and \ref{sec:unsupervised} provide an overview of supervised and unsupervised hyperbolic visual learning literature. Lastly in Section \ref{sec:conclusions} we outline advantages and improvements reported in current papers, as well as open challenges for the field.
\section{Background on hyperbolic geometry}
\label{sec:background}

\subsection{What is hyperbolic geometry?}
Hyperbolic geometry was initially developed in the 19th century by Gauss, Lobachevsky, Bolyai and others as a concrete example of a non-Euclidean geometry. Soon after it found important applications in physics, as the mathematical basis of Einstein's special theory of relativity. It can be characterized as the geometry of \emph{constant negative curvature}, differentiating it from the flat geometry of Euclidean space and the positively curved geometry of spheres and hyperspheres. From the point of view of representation learning, its attractive properties are its exponential expansion and its hierarchical, tree-like structure. Exponential expansion means that the volume of a ball in hyperbolic space growths exponentially with its diameter, in contrast to Euclidean space, where the rate of growth is polynomial. The `tree-likeness' of a metric space can be quantified by Gromov's hyperbolicity \citep{bridson2013metric}, which is zero for tree graphs, finite (but non-zero) for hyperbolic space, and infinite for Euclidean space. 

\subsection{Models of hyperbolic geometry}
Several different, but eventually equivalent, models of hyperbolic geometry exist \citep{cannon1997hyperbolic}. They differ in their coordinate representations of points and in their expressions for distances, geodesics, and other quantities. Although they can be converted into each other, certain models may be preferred for a given task, for reasons of numerical efficiency, ease of visualization, or simplified calculations. The most commonly used models are the Poincar\'e model, the hyperboloid (or `Lorentz') model,  the Klein model, and the upper half-space model.

\begin{itemize}
\item[$\bullet$] The \textbf{Poincar\'e model} represents $d$-dimensional hyperbolic space by the unit ball 
\[\mathbb{D}_d = \{p \in \mathbb{R}^d: p_1^2 + \dotsm + p_d^2 < 1\}\]
which, in the frequently considered case $d = 2$ becomes the unit disc. Geodesics (`shortest paths') are arcs of Euclidean circles (or lines), meeting the boundary of $\mathbb{D}_d$ at a right angle. While distances, area and volume are distorted in comparison to their Euclidean counterparts, the model is \emph{conformal}, i.e., hyperbolic angles are measured as in Euclidean geometry. In its two-dimensional form as Poincar\'e disc, the model is popular for visualizations; it is also the geometric basis of the art works Circle Limits I-IV of M.\ C.\ Escher; see Figure~\ref{fig:escher}.

\begin{figure}[t]
    \centering
    \includegraphics[width=0.8 \linewidth]{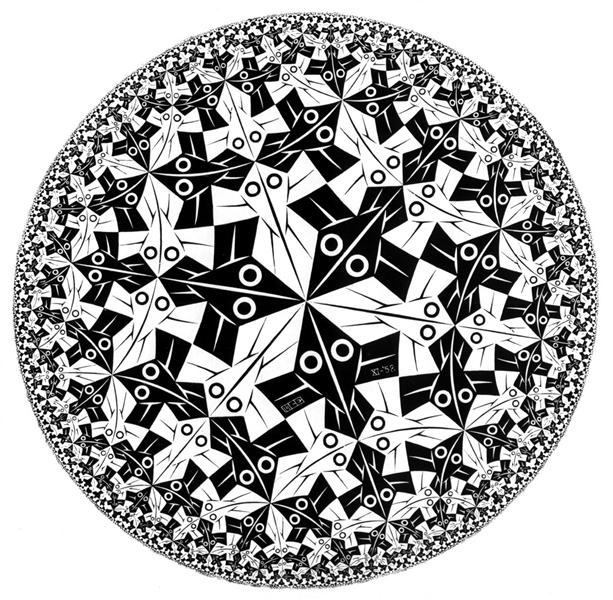}
    \caption{\textbf{Circle Limit I (1958).} This artwork by M.\ C.\ Escher is based on the Poincar\'e disc model of hyperbolic geometry.
    \label{fig:escher}}

\end{figure}

\item[$\bullet$] The \textbf{hyperboloid model} uses the single-sheet hyperboloid 
\[\mathbb{H}_d = \{x \in \mathbb{R}^{d+1}: x_0^2 - \left(x_1^2 + \dotsm + x_d^2\right) = 1,  x_0 > 0\}\]
as a model of $d$-dimensional hyperbolic geometry. Contrary to the other models, its ambient space $\mathbb{R}^{d+1}$ adds one dimension to the modeled space. Many formulas involving the hyperboloid model can be written in concise form by introducing the \emph{Lorentz product} $x \circ y = x_0 y_0 - (x_1 y_1 + \dotsm + x_d y_d)$. An advantage of the hyperboloid model is that it retains some linear structure;  translations and other isometries, for example, can be represented by linear maps. Expressions for distances and geodesics are simpler compared to other models. Notably, the Poincar\'e model can be derived as a projection (`stereographic projection') of the hyperboloid model to the unit ball \citep{cannon1997hyperbolic, ratcliffe1994foundations}.

\item[$\bullet$] The \textbf{Klein model} $\mathbb{K}_d$ also uses the unit ball to represent hyperbolic space. In contrast to the Poincar\'e model, it is not conformal; its geodesics, however, are Euclidean (`straight') lines, which can be beneficial from a computational point of view, \eg when computing barycenters. 

\item[$\bullet$] Lastly, the \textbf{upper half space model} represents $d$-dimensional hyperbolic space by the set $\mathbb{U}_d = \{x \in \mathbb{R}^d: x_d > 0\}$. It is a conformal model and shares many properties with the Poincar\'e model; geodesics, for example, are also arcs of Euclidean circles (or lines), meeting the boundary of $\mathbb{U}_d$ at a right angle. 
\end{itemize}

\subsection{Five core hyperbolic operations}
Within the context of deep learning and computer vision, we find that five core operations form the basic building blocks of the vast majority of algorithms. The ability to work with these five operations will cover most of existing literature:
\begin{enumerate}
\item Measuring the \textbf{distance} of two points $x$ and $y$;
\item Finding the \textbf{geodesic arc} (the distance-minimizing curve) from $x$ to $y$;
\item Forming a \textbf{geodesic}, by extending a geodesic arc as far as possible;
\item Using the \textbf{exponential map}, to determine the result of following a geodesic in direction $u$, at speed $r$, starting at a point $x$;
\item Moving a cloud of points, while preserving all their pairwise hyperbolic distances, by applying a \textbf{hyperbolic translation}.
\end{enumerate}
The distance of two points is given, in the Poincar\'e and the hyperboloid model respectively, by
\begin{align}
d_{\mathbb{D}}(p,q) &= \frac{1}{\sqrt{\kappa}} \arcosh\left(1 + \frac{2 |p - q|^2}{(1 - |p|^2)(1 - |q|^2)}\right),\\
d_{\mathbb{H}}(x,y) &= \frac{1}{\sqrt{\kappa}} \arcosh\left(x \circ y\right).
\end{align}
In the less frequently used Klein and the upper half space model, distances are given by
\begin{align}
d_{\mathbb{K}}(p,q) &= \frac{1}{\sqrt{\kappa}} \arcosh\left(\frac{1 - p^\top q}{\sqrt{1 - |p|^2}\sqrt{1 - |q|^2}}\right),\\
d_{\mathbb{U}}(x,y) &= \frac{1}{\sqrt{\kappa}} \arcosh\left(1 + \frac{|x-y|^2}{2x_d y_d} \right),
\end{align}
see \cite[\S6.1]{ratcliffe1994foundations}.
The scaling factor of distances is controlled by the \emph{curvature parameter} $\kappa \in (0,\infty)$, which is often standardized to $\kappa = 1$. The sectional curvature (in the sense of differential geometry) of hyperbolic space is constant, negative and equal to $-\kappa$. Given the distance function, it makes sense to speak of geodesics and geodesic arcs, that is (locally) distance-minimizing curves, either extending infinitely or connecting two points. In the hyperboloid model for example, each geodesic is the intersection of $\mathbb{H}_d$ with a Euclidean hyperplane in the ambient space $\mathbb{R}^{d+1}$. The geodesic at a point $x \in \mathbb{H}_d$ in direction $v$ can be written as
\begin{equation}\label{eq:geodesic_hyperboloid}
\lambda_{\mathbb{H}}(t) = \cosh(t\sqrt{\kappa})x + \sinh(t \sqrt{\kappa})u, \quad t \in \mathbb{R}.
\end{equation}
where $u$ is an element of the \emph{tangent space} $T_x = \{u\in \mathbb{R}^{d+1}: x \circ u = 0\}$, normalized to $u \circ u  = -1$. In the Poincar\'e model, the geodesics are precisely the segments of Euclidean circles and lines that meet the boundary of $\mathbb{D}_d$ at a right angle. 
A convenient formula for the geodesic arc between two points $p,q \in \mathbb{D}_d$ can be given in terms of gyrovectorspace calculus, see \eqref{eq:geodesic_poincare}.\\
The value of the exponential map $\exp_x(t u)$ is the result of following a geodesic in a normalized direction $u$ at a speed $t > 0$, after starting at a given point $x$ in hyperbolic space. Identifying $\mathbb{R}^d$ with the tangent space $T_x$ at $x$, the exponential mapping provides a convenient way to embed $\mathbb{R}^d$ into hyperbolic space with origin at $x$. The exponential map is the most often used function in hyperbolic learning for computer vision, as it allows us to map visual representations from Euclidean to hyperbolic space. In the hyperboloid model, the exponential mapping coincides with the expression of the geodesic given in \eqref{eq:geodesic_hyperboloid}.  In the Poincar\'e model the exponential map can be conveniently written in terms of gyrovectorspace addition and is given in \eqref{eq:exp_poincare}.

Finally, the hyperbolic translation $\tau_x$, also called Lorentz boost, Möbius transformation, or gyrovectorspace addition, is the unique distance-preserving transformation of hyperbolic space, which moves $0$ to a given point $x$. In the hyperboloid model, it can be represented by the linear map
\begin{align}\tau_x(y) &= L_x \cdot y, \quad \text{where}\\
L_x &= \begin{pmatrix}x_0 & \bar x^\top \\ \bar x & \sqrt{I_d + \bar x \bar x^\top}\end{pmatrix} \text{ with } \bar x = (x_0, \dotsc, x_d).
\end{align}
In the Poincar\'e model hyperbolic translations are also known as gyrovectorspace addition and form the basic operation of gyrovectorspace calculus.

\subsection{Gyrovectorspace calculus}
Gyrovectorspace calculus, as introduced by \cite{ungar2005gyrovector, ungar2012beyond}, provides a convenient and rapidly adopted framework for calculations in the Poincar\'e ball model. Its first basic operation is the (non-commutative) gyrovectorspace addition
\[p \oplus q = \frac{(1 - |p|^2) q + (1 + 2p^\top q + |q|^2)p}{1 + 2p^\top q + |p|^2|q|^2}.\]
As a secondary operation, the (commutative) gyrovectorspace scalar product
\[t \otimes p = p \otimes t = \tanh\big(t \artanh(|p|)\big) \frac{p}{|p|}\]
with a scalar $t \in \mathbb{R}$ is introduced. Hyperbolic translations are directly given by $\tau_p(q) = p \oplus q$
and the geodesic arc connecting $p$ and $q$ is
\begin{equation}\label{eq:geodesic_poincare}
\lambda_{\mathbb{D}}(t) = p \oplus \Big(\big((-p) \oplus q\big) \otimes t \Big), \quad t \in [0,1].
\end{equation}
Letting $t$ range through all of $\mathbb{R}$ a full geodesic line is obtained.

In the context of gyrovector space calculus, the Poincar\'e ball is often rescaled with the square root of curvature, setting
\[\mathbb{D}^d_\kappa = \{p \in \mathbb{R}^d: p_1^2 + \dotsm + p_d^2 < 1/\kappa\}.\]
The advantage of this rescaling is that Euclidean space is obtained as a continuous limit as $\kappa \to 0$. 
In the rescaled model, gyrovectorspace addition and scalar product become
\[p \oplus_\kappa q = \tfrac{1}{\sqrt{\kappa}} \left((\sqrt{\kappa} p) \oplus (\sqrt{\kappa} q)\right)\]
and
\[t \otimes_\kappa p = \tfrac{1}{\sqrt{\kappa}} (t \otimes (\sqrt{\kappa} p))\]
for $p,q \in \mathbb{D}^d_\kappa$. The exponential map in direction of a tangent vector $v \in T_p$ can then be written as 
\begin{equation}\label{eq:exp_poincare}
\exp_x^\kappa(v) = x \oplus_\kappa \left(\tanh\left(\frac{\sqrt{\kappa}|v|}{1 - \kappa |x|^2}\right) \frac{v}{\sqrt{\kappa}|v|}\right)
\end{equation}
for $p \in \mathbb{D}^d_\kappa$, see \cite{ganea2018hyperbolic}.

\subsection{Non-visual hyperbolic learning}
The traction of hyperbolic learning in computer vision is built upon advances in embedding hierarchical structures, designing hyperbolic network layers, and hyperbolic learning on other data types such as graphs, text, and more. Below, we discuss these works and their relevance for hyperbolic visual learning literature.

\paragraph{Hyperbolic embedding of hierarchies.}
Embedding hierarchical structures like trees and taxonomies in Euclidean space suffers from large distortion~\citep{bachmann2020constant}, and polynomial volume expansion, limiting its capacity to capture the exponential complexity of hierarchies. However, hyperbolic space can be thought of as a continuous version of trees~\citep{nickel2017poincare} and has tree-like properties~\citep{hamann2018tree, ungar2008gyrovector}, like the exponential growth of distances when moving from the origin towards the boundary.
Encouraged by this,~\citet{nickel2017poincare} propose to embed hierarchical structures on the Poincar\'e model. The goal is to learn hyperbolic representations for the nodes of a hierarchy, such that the distance in the embedding space has an inverse relation with semantic similarity. Let $\mathcal{D} = \{(u, v)\}$ denote the set of the nodes connected in a given hierarchy. To embed the nodes in the Poincar\'e model, ~\citet{nickel2017poincare} minimize the following loss function:
\begin{equation}
\mathcal{L}(\Theta) = \sum_{(u,v) \in \mathcal{D}} \log \frac{e^{-d(u,v)}}{\sum_{v^{'} \in \mathcal{N}(u)} e^{-d(u,v^{'})}},
\end{equation}
where $\mathcal{N}(u) = \{v^{'}|(u,v^{'}) \notin \mathcal{D}\} \cup \{v\}$ denotes the set of the nodes not related to $u$, including $v$, as negative examples. The loss function pushes unrelated nodes farther apart than the related ones. To evaluate the embedded hierarchy, the distances between pairs of connected nodes $(u,v)$ are calculated and ranked among the negative pairs of nodes (\ie the nodes not in $\mathcal{D}$), and the mean average precision (MAP) is calculated based on the ranking. Later,~\citet{sala2018representation} propose a combinatorial construction to embed the trees in hyperbolic space without optimization and with low distortion, relieving the optimization problems in existing works.
~\citet{ganea2018hyperbolic2} address drawbacks of~\citep{nickel2017poincare} including the collapse of the points on the boundary of the space as a result of the loss function and incapability of encoding asymmetric relations. They introduce entailment cones to embed hierarchies, using a max-margin loss function:
\begin{equation}
    \mathcal{L} = \sum_{(u,v) \in \mathcal{P}} E(u,v) + \sum_{(u^{'}, v^{'}) \in \mathcal{N}} \max(0, \gamma - E(u^{'}, v^{'})),
\end{equation}
where $\gamma$, $\mathcal{P}$, and $\mathcal{N}$ indicate margin, the positive and negative edges, respectively. $E(u,v)$ is a penalty term that forces child nodes to fall under the cone of the parent node.
Amongst others, hyperbolic embeddings have been proposed for multi-relational graphs~\citep{balazevic2019multi}, low-dimensional knowledge graphs~\citep{chami2020low}, and learning continuous hierarchies in Lorentz model~\citep{nickel2018learning}.

\paragraph{Hyperbolic neural networks.} Foundational in the transition of deep learning towards hyperbolic space is the development of hyperbolic network layers and their optimization. We consider two pivotal papers here that provide a such theoretical foundation, namely Hyperbolic Neural Networks by \cite{ganea2018hyperbolic} and Hyperbolic Neural Networks++ by \cite{shimizu2021hyperbolic}.
~\cite{ganea2018hyperbolic} propose multinomial logistic regression in the Poincar\'e ball. 
Given $k \in \{1, ..., K\}$ classes, $p_k \in \mathbb{D}^n_c$, and $a_k \in \mathbb{D}^n_c \setminus \{0\}$, hyperbolic logistic regression is performed using 
\begin{equation}
 \begin{split}
    p(y=k|x) \propto
    & \exp(\frac{\lambda^c_{pk}\lVert a_k \rVert}{\sqrt{c}}\\
    &\sinh^{-1}(\frac{2\sqrt{c}\langle -p_k \oplus _c x, a_k \rangle}{(1-c\lVert -p_k\oplus _c x \rVert^2)\lVert a_k \rVert})).
\end{split}
\label{eq:ganea-mlr}
\end{equation}
As an extension, a hyperbolic version of linear layer $f$
is given as $f : \mathbb{R}^n \rightarrow \mathbb{R}^m$, a Möbius version of $f$ where the map from $\mathbb{D}^n \rightarrow \mathbb{D}^m$ is defined as:
\begin{equation}
    f^{\otimes_c} \coloneqq \exp ^c _0(f(\log^c_0(x))),
\end{equation}
with $\exp ^c _0 : T_{0_m}\mathbb{D}^m_c \rightarrow \mathbb{D}^m_c$ and $\log^c_0: \mathbb{D}^n_c \rightarrow T_{0_n}\mathbb{D}^n_c$. They furthermore outline how to create recurrent network layers.

~\cite{shimizu2021hyperbolic} reformulate the hyperbolic logistic regression of \citep{ganea2018hyperbolic} to reduce the number of parameters to the same level as the Euclidean logistic regression. The new formulation is $p(y=k|x) \propto \exp(v_k(x))$, where
\begin{equation}
\begin{split}
    v_k(x) =& \ 2c^{-\frac{1}{2}}\lVert z_k \rVert\sinh^{-1}(\lambda^c_{x}\langle\sqrt{c}x, [z_k]\rangle\cosh(2\sqrt{c}r_k)\\
    &- (\lambda^c_{x} - 1)\sinh(2\sqrt{c}r_k))
\end{split}
\end{equation}
where $r_k \in \mathbb{R}$ and $z_k\in T_0\mathbb{B}^n_c = \mathbb{R}^n$ are the parameters for each class. In turn, their linear layer is given as
\begin{equation}
    y=\mathcal{F}^c(x;Z,r)\coloneqq w(1+\sqrt{{1+c\lVert w\rVert^2}})^{-1}
\end{equation}
where $Z=\{z_k\in T_0\mathbb{B}^n_c = \mathbb{R}^n\}^m_{k=1}$, $r=\{r_k \in \mathbb{R}\}^m_{k=1}$, and $w \coloneqq(c^{-\frac{1}{2}}\sinh(\sqrt{c}v_k(x)))^m_{k=1}$. More importantly for computer vision, they show how to formulate convolutional layers using Poincar\'e fully connected layer and $\beta$-concatenation. To do so, they show how to generalize the hyperbolic linear layer to image patches through $\beta$-splits, and $\beta$-concatenation, leading in principle to arbitrary-dimensional convolutional layers. Moreover, Poincar\'e multi-head attention is possible through the same operators.

\begin{figure*}[t]
\centering
\includegraphics[width=\textwidth]{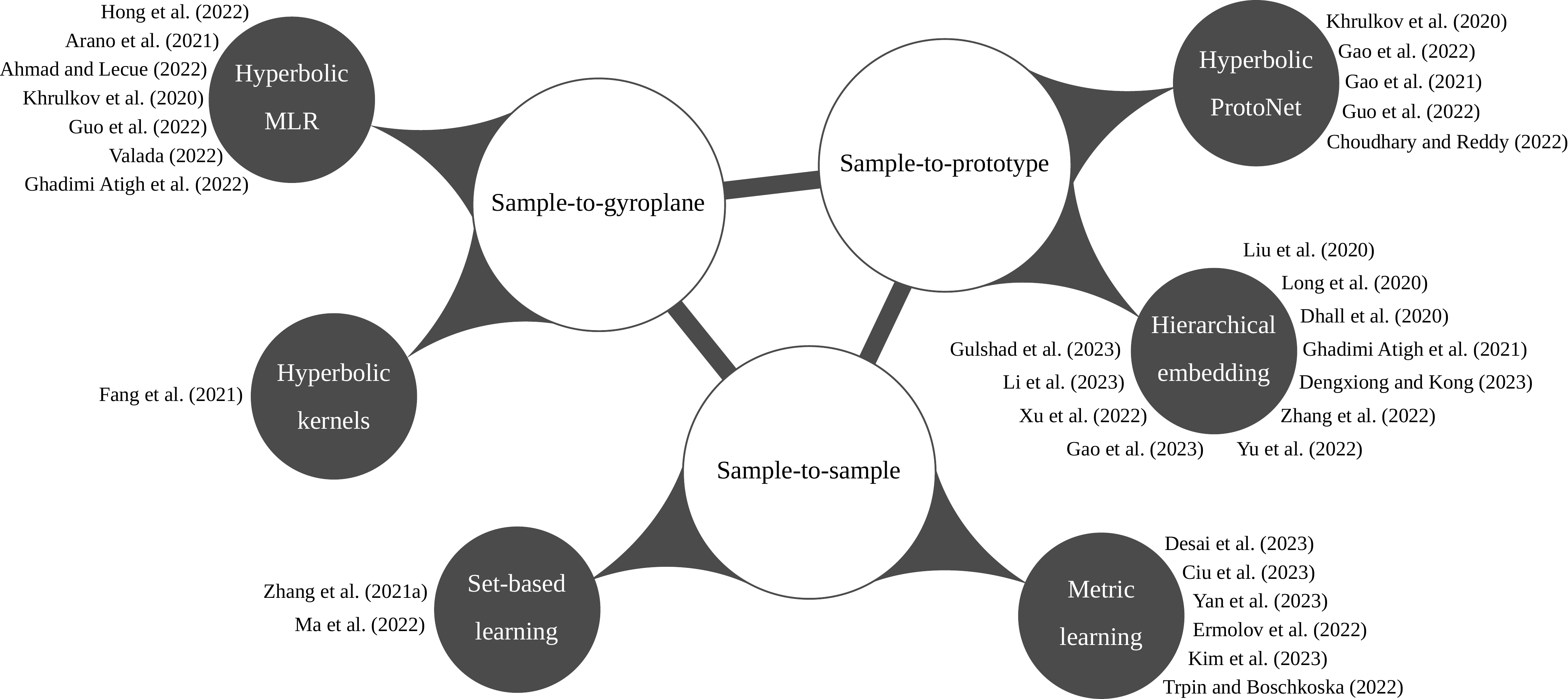}
\caption{\textbf{The three core strategies for supervised hyperbolic learning in computer vision.} Current literature performs hyperbolic learning of visual embeddings by learning to match training samples (i) to hyperbolic class hyperplanes, \ie gyroplanes, (ii) to hyperbolic class prototypes, or (iii) by contrasting to other samples.}
\label{fig:supervised-venn}
\end{figure*}.

\paragraph{Hyperbolic learning of graphs, text, and more.}
The advances in hyperbolic embeddings of hierarchies and the introduction of hyperbolic network layers have spurred research in several other research directions as well. As a logical extension of hierarchical embeddings, graph networks have been extended to hyperbolic space. \citet{liu2019hyperbolic} and \citet{chami2019hyperbolic} propose a tangent-based view to hyperbolic graph networks. Both approaches model a graph layer by first mapping node embeddings to the tangent space, then performing the transformation and aggregation in the tangent space, after which the updated node embeddings are projected back to the hyperbolic manifold at hand. Since tangent operations only provide an approximation of the graph operations on the manifold, several works have proposed graph networks that better abide the underlying hyperbolic geometry, such as constant curvature $\kappa$-GCNs \citep{bachmann2020constant}, hyperbolic-to-hyperbolic GCNs \citep{dai2021hyperbolic}, Lorentzian GCNs \citep{zhang2021lorentzian}, and attention-based hyperbolic graph networks \citep{gulcehre2019hyperbolic,zhang2021hyperbolic}. Hyperbolic graph networks have shown to improve node, link, and graph classification compared to Euclidean variants, especially when graphs have latent hierarchical structures.

Hyperbolic embeddings have also been investigated for text. \cite{tifrea2019poincar}, \citet{dhingra2018embedding}, and \citet{leimeister2018skip} propose hyperbolic alternatives for word embeddings. \cite{zhu2020hypertext} introduce HyperText to endow FastText with hyperbolic geometry. Embedding text in hyperbolic space has the potential to improve similarity, analogy, and hypernymy detection, most notably with few embedding dimensions.

Beyond text and graphs, hyperbolic learning has shown to be beneficial for several other research directions, including but not limited to learning representations for molecular/cellular structures \citep{klimovskaia2020poincare,yu2020semi,wu2021hyperbolic}, recommender systems \citep{mirvakhabova2020performance,wang2021fully,yang2022hrcf}, skeletal action recognition \citep{franco2023hyperbolic}\lastpaper{, LiDAR data \citep{tong2022hyperbolic,wang2023hypliloc}, point clouds \citep{montanaro2022rethinking, anvekar2023gpr}, and 3D shapes \citep{chen2020learning2}}. In summary, hyperbolic geometry has impacted a wide range of research fields. This survey focuses specifically on the impact and potential in the visual domain.
\section{Supervised hyperbolic visual learning}
\label{sec:supervised}
In Figure~\ref{fig:supervised-venn}, we provide an overview of literature on supervised learning with hyperbolic geometry in computer vision. In current vision works, hyperbolic learning is mostly performed at the embedding- or classifier-level. In other words, current works rely on standard networks for feature learning and transform the output embeddings to hyperbolic space for the final learning stage. For supervised learning in hyperbolic space, we have identified three main optimization strategies:
\begin{enumerate}
\item \textbf{Sample-to-gyroplane learning} denotes the setting where classes are represented by hyperbolic hyperplanes, \ie gyroplanes, with networks optimized based on confidence logit scores between samples and gyroplanes.
\item \textbf{Sample-to-prototype learning} denotes the setting where class semantics are represented as points in hyperbolic space, and networks are optimized to minimize hyperbolic distances between samples and prototypes.
\item \textbf{Sample-to-sample learning} denotes the setting where networks are optimized by learning metrics or contrastive objectives between samples in a batch.
\end{enumerate}
For all strategies, let $(x,y)$ denote the visual input $x$, which can be an image or a video, and the corresponding label $y \in \mathcal{Y}$. Let $f_\theta(x) \in \mathbb{R}^{D}$ denote its Euclidean embedding after going through a network. This representation is mapped to hyperbolic space using the exponential map, denoted as $g(x) = \exp_0(f_\theta(x))$. In many hyperbolic works, additional information about hierarchical relations between classes is assumed. Let $\mathcal{H} = (\mathcal{Y}, \mathcal{P}, \mathcal{R})$, with $\mathcal{Y}$ the class labels denoting the leaf nodes of the hierarchy, $\mathcal{P}$ the internal nodes, and $\mathcal{R}$ the set of hypernym-hyponym relations of the hierarchy. Below, we discuss how current literature tackles each strategy in detail sequentially.

\subsection{Sample-to-gyroplane learning}
The most direct way to induce hyperbolic geometry in the classification space is by replacing the classification layer by a hyperbolic alternative. This can be done either by means of a hyperbolic logistic regression or through hyperbolic kernel machines.

\paragraph{Hyperbolic logistic regression.} \cite{khrulkov2020hyperbolic} incorporate a hyperbolic classifier by taking a standard convolutional network and mapping the outputs of the last hidden layer to hyperbolic space using an exponential map. Afterwards, the hyperbolic multinomial logistic regression as described by \cite{ganea2018hyperbolic} is used to obtain class logits which can be optimized with cross-entropy. They find that training a hyperbolic classifier on top of a convolutional network allows us to obtain uncertainty information based on the distance to the origin of the hyperbolic embeddings of images. Out-of-distribution samples on average have a smaller norm, making it possible by differentiating in- to out-of-distribution samples by sorting them by the distance to the origin. \newpapertwo{~\cite{hong2022curved} show that hyperbolic classification is beneficial for visual anomaly recognition tasks, such as out-of-distribution detection in image classification and segmentation tasks.~\cite{arano2021multimodal} use hyperbolic layers to perform multi-modal sentiment analysis based on the audio, video, and text modalities.~\cite{ahmad2022fisheyehdk} also show the effect of hyperbolic space to perform object recognition with ultra-wide field-of-view lenses.} 

\cite{guo2022clipped} address a limitation when training classifiers in hyperbolic space, namely a vanishing gradient problem due to the hybrid architecture of current hyperbolic approaches in computer vision, where Euclidean features are connected to a hyperbolic classifier. Equation \ref{eq:ganea-mlr} highlights that to maximize the likelihood of correct predictions, the distance to hyperbolic gyroplanes needs to be maximized. In practice, embeddings of samples are pushed to the boundary of the Poincar\'e ball. As a result, the inverse of the Riemannian tensor metric approaches zero, resulting in small gradients. This finding is in line with several other works on vanishing gradients in hyperbolic representation learning \citep{nickel2018learning,
liu2019hyperbolic}.

To combat the vanishing gradient problem, \cite{guo2022clipped} propose to clip the Euclidean embeddings of samples before the exponential mapping, \ie:
\begin{equation}
f^{\text{clipped}}_\theta(x) = \min \big \{ 1, \frac{r}{||f_\theta(x)||} \big \} \cdot f_\theta(x),
\end{equation}
with $r$ as a hyperparameter. This trick improves learning with hyperbolic multinomial logistic regression, especially when dealing with many classes such as on ImageNet. Furthermore, training with clipped hyperbolic classifiers improves out-of-distribution detection over training with Euclidean classifiers, while also being more robust to adversarial attacks.

\begin{figure}[t]
    \centering
    \includegraphics[width=\linewidth]{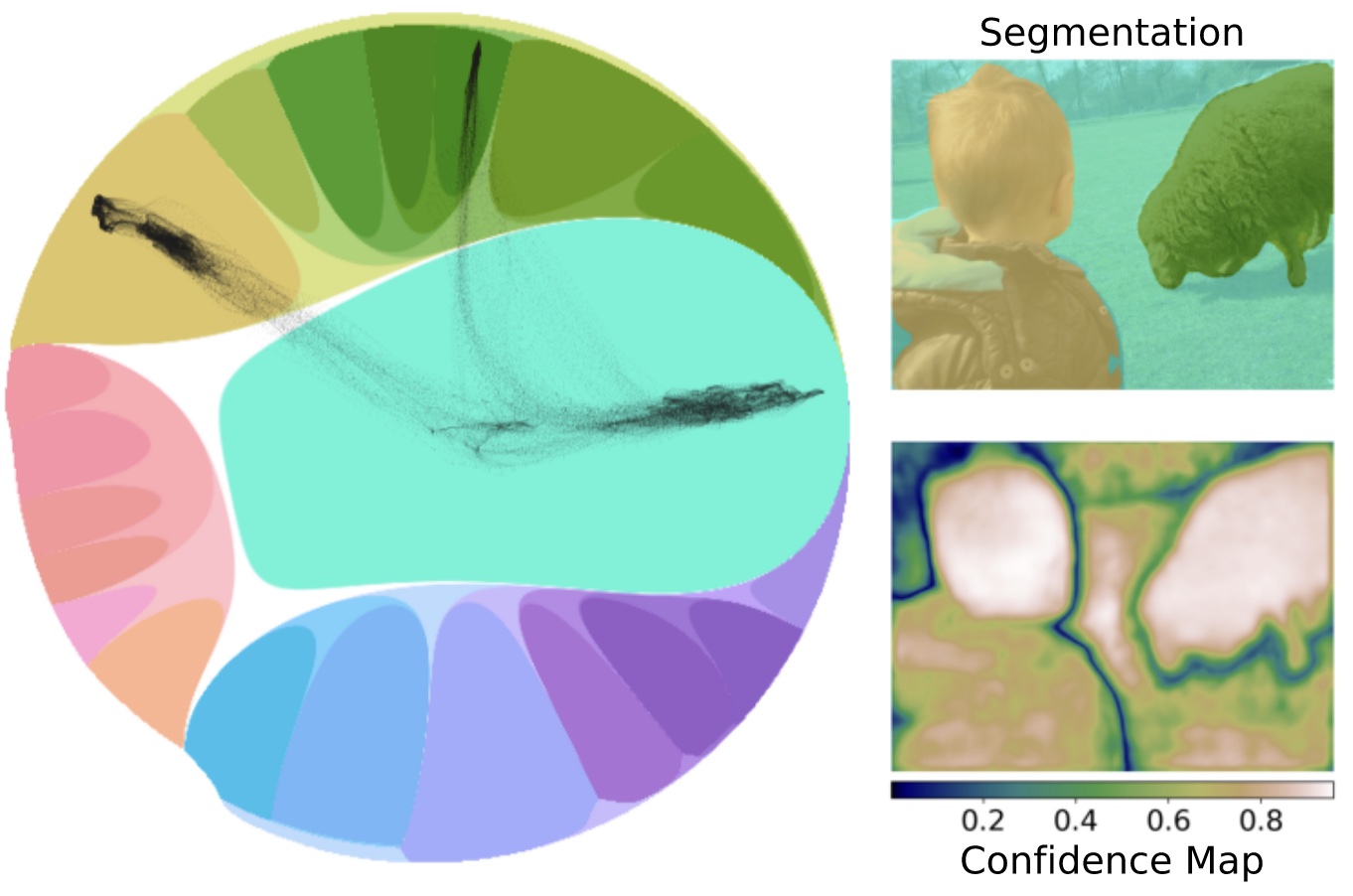}
    \caption{\textbf{Hyperbolic image segmentation} naturally provides us per-pixel uncertainty information. Pixels with low hyperbolic norm constitute pixels with high uncertainty and are strongly correlated with closeness to semantic boundaries. Image courtesy of \cite{atigh2022hyperbolic}.}
    \label{fig:supervised-segmentation}
\end{figure}

Next to global classification, a few recent works have investigated hyperbolic logistic regression for structured prediction tasks such as object detection and image segmentation. \cite{valada2022hyperbolic} extend object detection with hyperbolic geometry, amongst others by replacing the classifier head of a two-stage detection like Sparse R-CNN \citep{sun2021sparse} with a hyperbolic logistic regression, improving object detection performance in standard and zero-shot settings.
\cite{atigh2022hyperbolic} introduce Hyperbolic Image Segmentation, where the final per-pixel classification was performed in hyperbolic space. Starting from the geometric interpretation of hyperbolic gyroplanes of \cite{ganea2018hyperbolic}, they find that simultaneously computing class logits over all pixels of all images in a batch, as is customary in Euclidean networks, is not directly applicable in hyperbolic space. This is because the explicit computation of the M\"obius addition requires evaluating a tensor in $\mathbb{R}^{W \times H \times |\mathcal{Y}| \times d}$ for an images of size $(W \times H)$ with $d$ embedding dimensions. Instead, they rewrite the M\"obius addition as:
\begin{equation}
\begin{split}
f_1 \oplus_c f_2 = & \alpha f_1 + \beta f_2,\\
\alpha = & \frac{1 + 2c \langle f_1, f_2 \rangle + c ||f_2||^2}{1 + 2c \langle f_1, f_2 \rangle + c^2 ||f_1||^2||f_2||^2},\\
\beta = & \frac{1 + c||f_1||^2}{1 + 2c \langle f_1, f_2 \rangle + c^2 ||f_1||^2||f_2||^2}.
\end{split}
\end{equation}
This rewrite reduces the addition to adding two tensors in $\mathbb{R}^{W \times H \times |\mathcal{Y}|}$, allowing for per-pixel evaluation on image batches. For training, \cite{atigh2022hyperbolic} incorporate hierarchical information by replacing the one-hot softmax with a hierarchical softmax:
\begin{equation}
p(\hat{y} = y | g(x)_{ij}) = \prod_{h \in \mathcal{H}_y} \frac{\exp(\xi_h(g(x)_{ij}))}{\sum_s \in S_h \exp(\xi_s(g(x)_{ij}))},
\end{equation}
with $\mathcal{H}_y = \{y\} \cap \mathcal{A}_y$ the set containing $y$ and its ancestors and $S_h$ the set of siblings of class $h$. Performing per-pixel classification with hyperbolic hierarchical logistic regression opens up multiple new doors for image segmentation. First, the notion of uncertainty as given by the hyperbolic norm of output embeddings generalizes naturally to the pixel level. As shown in Figure~\ref{fig:supervised-segmentation}, the norm of pixel embeddings correlates with semantic ambiguity; the closer the pixel is to a semantic boundary the lower the pixel norm. \cite{chen2022hyperbolic} have already used this insight to improve image segmentation. They outline a hyperbolic uncertainty loss, where the cross-entropy loss of a pixel is weighted as follows for pixel $x_{ij}$:
\begin{equation}
\text{uw}(x_{ij}) = 1 + \frac{1}{\log(t + \frac{d_h(g(x)_{ij}, 0)}{d_h(g(s), 0)})},
\end{equation}
with $s$ the most confident pixel and $t$ a hyperparameter set to 1.02 in order to have a wide weight variation while avoiding division by zero. Adding this weight to the cross-entropy pixel loss consistently improves segmentation results for well-known segmentation networks. Other benefits of hyperbolic image segmentation include better zero-label generalization and higher effectiveness with few embedding dimensions compared to Euclidean pixel embeddings.

\begin{figure*}
\centering
\includegraphics[width=\textwidth]{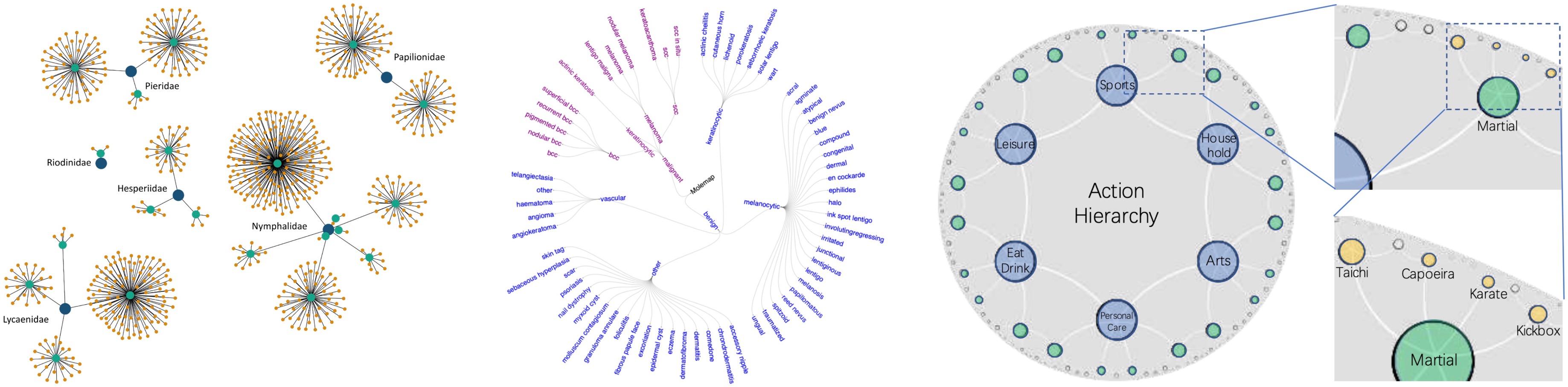}
\caption{\textbf{Hierarchical knowledge amongst classes provides a structure for hyperbolic embeddings} in computer vision approaches, where classes are represented as points or prototypes in hyperbolic space according to their hypernym-hyponym relations. For example, \cite{dhall2020hierarchical} exploit hierarchical relations from entomological collections (left), while \cite{yu2022skin} utilize taxonomies of skin lesion diseases (middle) and \cite{long2020searching} do the same for action hierarchies (right). Images courtesy of the respective publications.}
\label{fig:hierarchies}
\end{figure*}

\paragraph{Hyperbolic kernel machines.} Next to logistic regression, \cite{cho2019large} provide a general formulation for kernel methods in hyperbolic space with large-margin classifiers. \cite{fang2021kernel} introduce positive definite kernel functions in hyperbolic space and show its potential for computer vision. Specifically, they propose hyperbolic instantiations of tangent kernels, radial basis function kernels, (generalized) Laplace kernels, and binomial kernels. The kernels can be plugged on top of convolutional networks and trained with cross-entropy to benefit from both the representation learning of the convolutional layers and the hyperbolic kernel dynamics in the classifier. Deep learning with hyperbolic kernel methods improves few-shot learning, person re-identification, and knowledge distillation. Zero-shot learning is even enabled through kernel distances between visual embeddings and semantic class representations.

\subsection{Sample-to-prototype learning}
The most popular strategy in hyperbolic learning is to represent classes as prototypes, \ie as points in hyperbolic space. In this research direction, there are two solutions: embedding classes based on their sample mean, in the spirit of Prototypical Networks (ProtoNet) \citep{snell2017prototypical}, or embeddings classes based on a given hierarchy over all classes.

\paragraph{Hyperbolic ProtoNet.}
In Prototypical Networks \citep{snell2017prototypical}, the prototype of a class $k$ is determined as the mean vector of the samples belonging to that class:
\begin{equation}
P_{\mathbb{R}}(k) = \frac{1}{|S_k|} \sum_{y_s \in S_k} f_\theta(x_s),
\end{equation}
with $S_k$ the set of samples belonging to class $k$. Inference can in turn be performed by assigning the label of the nearest prototype for a test sample. \citet{khrulkov2020hyperbolic} generalize this formulation to Hyperbolic Prototypical Networks. Since computing averages in the Poincar\'e ball model requires expensive Fr\'echet mean calculations, they perform averaging using the Einstein midpoint, given in Klein coordinates as:
\begin{equation}
P_{\mathbb{K}}(k) = \sum_{i=1}^{|S_k|} \gamma_i g_{\mathbb{K}}(x_i) / \sum_{i=1}^{|S_k|} \gamma_i,
\end{equation}
with $\gamma_i$ the Lorentz factors:
\begin{equation}
\gamma_i = \frac{1}{\sqrt{1 - c||g(x_i)||^2}}.
\end{equation}
Since \citet{khrulkov2020hyperbolic} operate in the Poincar\'e ball model, this averaging operation requires transforming embeddings to and from the Klein model:
\begin{equation}
\begin{split}
g_{\mathbb{K}}(x_i) = & \frac{2 g_{\mathbb{D}}(x_i)}{1 + c||g_{\mathbb{D}}(x_i)||^2},\\
g_{\mathbb{D}}(x_i) = & \frac{g_{\mathbb{K}}(x_i)}{1 + \sqrt{1 - c||g_{\mathbb{K}}(x_i)||^2}},
\end{split}
\end{equation}
with $g_{\mathbb{D}}(x_i)$ and $g_{\mathbb{K}}(x_i)$ the embeddings of input $x_i$ in respectively the Poincar\'e ball model and the Klein model.
Akin to its Euclidean counterpart, Hyperbolic ProtoNet is used to address few-shot learning, where the sample mean prototype serves as the class representation. \cite{khrulkov2020hyperbolic} show that performing prototypical few-shot learning in hyperbolic space is competitive to Euclidean prototypical learning, even resulting in better accuracy scores when relying on a 4-layer ConvNet as the backbone.

As a follow-up work, \cite{gao2021curvature} show that different tasks and even individual classes in few-shot learning favor different curvatures. They propose to generate a per-class curvature based on the second-order statistics of its in-class and out-of-class sample representations. Using the second-order statistics, a multi-layer perceptron with sigmoid activation is learned to fix the range of the curvature to $[0,1]$. Given class-specific curvatures, prototypes are obtained by constructing an intra-class distance matrix on top of which an MLP is trained. The MLP serves as weights for each in-class sample. The procedure is repeated for the closest samples in the out-of-class set, after which the per-class prototype is given as the weighted hyperbolic average over the in-class and closest out-of-class samples. The curvature generation and weighted hyperbolic averaging improve few-shot learning in both inductive and transductive settings.

The hyperbolic clipping of \cite{guo2022clipped} is also effective for few-shot learning, consistently outperforming the standard ProtoNet and Hyperbolic ProtoNet on the CUB Birds and miniImageNet few-shot benchmarks. A few other works have extended Hyperbolic ProtoNet for few-shot learning with set- and grouplet-based learning and will be discussed in the sample-to-sample learning section.

\newpaper{Recently,~\cite{gao2022hyperbolic} investigate feature augmentation in hyperbolic space to solve the overfitting problem when dealing with limited data.
On top, they introduce a scheme to estimate the feature distribution using neural-ODE. These elements are then plugged into few-shot approaches such as the hyperbolic prototypical networks of \cite{khrulkov2020hyperbolic}, improving performance.
~\cite{choudhary2022towards} improve hyperbolic few-shot learning by reformulating hyperbolic neural networks through Taylor series expansions of hyperbolic trigonometric functions and show that it improves the scalability and compatibility, and outperforms Hyperbolic ProtoNet. }

\paragraph{Hierarchical embedding of prototypes.}
Where Hyperbolic ProtoNets are effective in few-shot settings, a number of works have also investigated prototype-based solutions for the general classification. As starting point, these works commonly assume that the classes in a dataset are organized in a hierarchy, see Figure~\ref{fig:hierarchies}. \cite{long2020searching} embed action class hierarchy $\mathcal{H}$ in hyperbolic space using hyperbolic entailment cones \citep{ganea2018hyperbolic2}, with an additional loss to increase the angular separation between leaf nodes to avoid inter-label confusion amongst class labels $\mathcal{Y}$. With $\mathcal{L}_H(\mathcal{H})$ as the hyperbolic embedding loss for hierarchy $\mathcal{H}$, let $\mathcal{P}$ denote the leave nodes of the hierarchy. Then the separation-based loss is given over the leaf nodes as:
\begin{equation}
\mathcal{L}_S(\mathcal{P}) = \mathbf{1}^T(\hat{P} \hat{P}^T - I) \mathbf{1},
\end{equation}
with $\hat{P}$ the $\ell_2$-normalized representations of the leaf nodes. By combining the hierarchical and separation based losses, the hierarchy is embedded to balance both hierarchical constraints and discriminative abilities. The embedding is learned \emph{a priori}, after which video embeddings are projected to the same hyperbolic space and optimized to their correct class embedding. This approach improves action recognition, zero-shot action classification, and hierarchical action search. In a similar spirit, \cite{dhall2020hierarchical} show that using hyperbolic entailment cones for image classification is empirically better than using Euclidean entailment cones. Rather than separating hierarchical and visual embedding learning, \cite{yu2022skin} propose to simultaneously learn hierarchical and visual representations for skin lesion recognition in images. Image embeddings are optimized towards their correct class prototype, while the classes are optimized to abide by their hyperbolic entailment cones with an extra distortion loss to obtain better hierarchical embeddings. \newpaper{\cite{gulshad2023hierarchical} propose Hierarchical Prototype Explainer, a reasoning model in hyperbolic space to provide explainability in video action recognition. Their approach learns hierarchical prototypes at different levels of granularity~\eg parent and grandparent levels, to explain the recognized action in the video. By learning the hierarchical prototypes, they can provide explanations on different levels of granularity, including interpretation of the prediction of a specific class label and providing information on the spatiotemporal parts that contribute to the final prediction.} \newpapertwo{~\cite{li2023isolated} investigate the semantic space of action recognition datasets and bridge the gap between different labeling systems. To achieve a unified action learning, actions are connected into a hierarchy using  VerbNet~\citep{schuler2005verbnet} and embedded as prototypes in hyperbolic space.}
 
Hierarchical prototype embeddings have also been successfully employed in the zero-shot domain. \cite{liu2020hyperbolic} show how to perform zero-shot learning with hyperbolic embeddings. Classes are embedded by taking their WordNet-based Poincar\'e Embeddings \citep{nickel2017poincare} and text-based Poincar\'e GloVe embeddings \citep{tifrea2019poincar}. Both are concatenated to obtain class prototypes. By optimizing seen training images to their prototypes, it becomes possible to generalize to unseen classes during testing through a nearest neighbor search in the concatenated hyperbolic space. \cite{xu2022meta} also perform hyperbolic zero-shot learning by training hyperbolic graph layers \citep{chami2019hyperbolic} on top of hyperbolic word embeddings.
\newpaper{\cite{dengxiong2023ancestor} show the potential of hyperbolic space in generalized open set recognition, which classifies unknown samples based on side information. A side information (taxonomy) learning framework is introduced to embed the information in hyperbolic space with low distortion and identify the unknown samples. Moreover, an ancestor search algorithm is outlined to find the most similar ancestor in the taxonomy of the known classes.}

For standard classification, \cite{ghadimi2021hyperbolic} show how to integrate uniformity amongst prototypes in hyperbolic space by embedding classes with maximum separation on the boundary of the Poincar\'e ball given by \citep{mettes2019hyperspherical,kasarla2022maximum}. With prototypes now at the boundary of the ball, standard distance functions no longer apply since they are at the infinite distance to any point within the ball. To that end, they propose to use the Busemann distance, which is given for hyperbolic image embedding $g(x)$ and prototype $p$ as:
\begin{equation}
b_{p}(g(x)) = \log( \frac{||p - g(x)||^2}{1 - ||g(x)||^2}).
\end{equation}
By fixing prototypes with maximum separation \emph{a priori} and minimizing this distance function with an extra regularization towards the origin, it becomes possible to perform hyperbolic prototypical learning with prototypes at the ideal boundary. \cite{ghadimi2021hyperbolic} show that such an approach has direct links with conventional logistic regression in the binary case, highlighting its inherent properties. Moreover, maximally separated prototypes can also be replaced by prototypes from word embeddings or hierarchical knowledge, depending on the available knowledge and task at hand. \newpapertwo{In addition to standard classification, hierarchical hyperbolic embeddings have demonstrated effectiveness in continual learning~\citep{gao2023exploring}. To learn the new data,~\cite{gao2023exploring} propose a dynamically expanding geometry through a mixed-curvature space, enabling learning of complex hierarchies in a data stream. To prevent forgetting, angle-regularization and neighbor-robustness losses are used to preserve the geometry of the old data.}

Few-shot learning has also been investigated with hierarchical knowledge. \cite{zhang2022hyperbolic} perform such few-shot learning by first training a network on a joint classification and hierarchical consistency objective. The classification is given as a softmax over the class probabilities, as well as the softmax over the superclasses. In the few-shot inference stage, class prototypes are obtained through hyperbolic graph propagation to deal with the limited sample setting, improving few-shot learning as a result.

\begin{figure}[t]
\centering
\includegraphics[width=\linewidth]{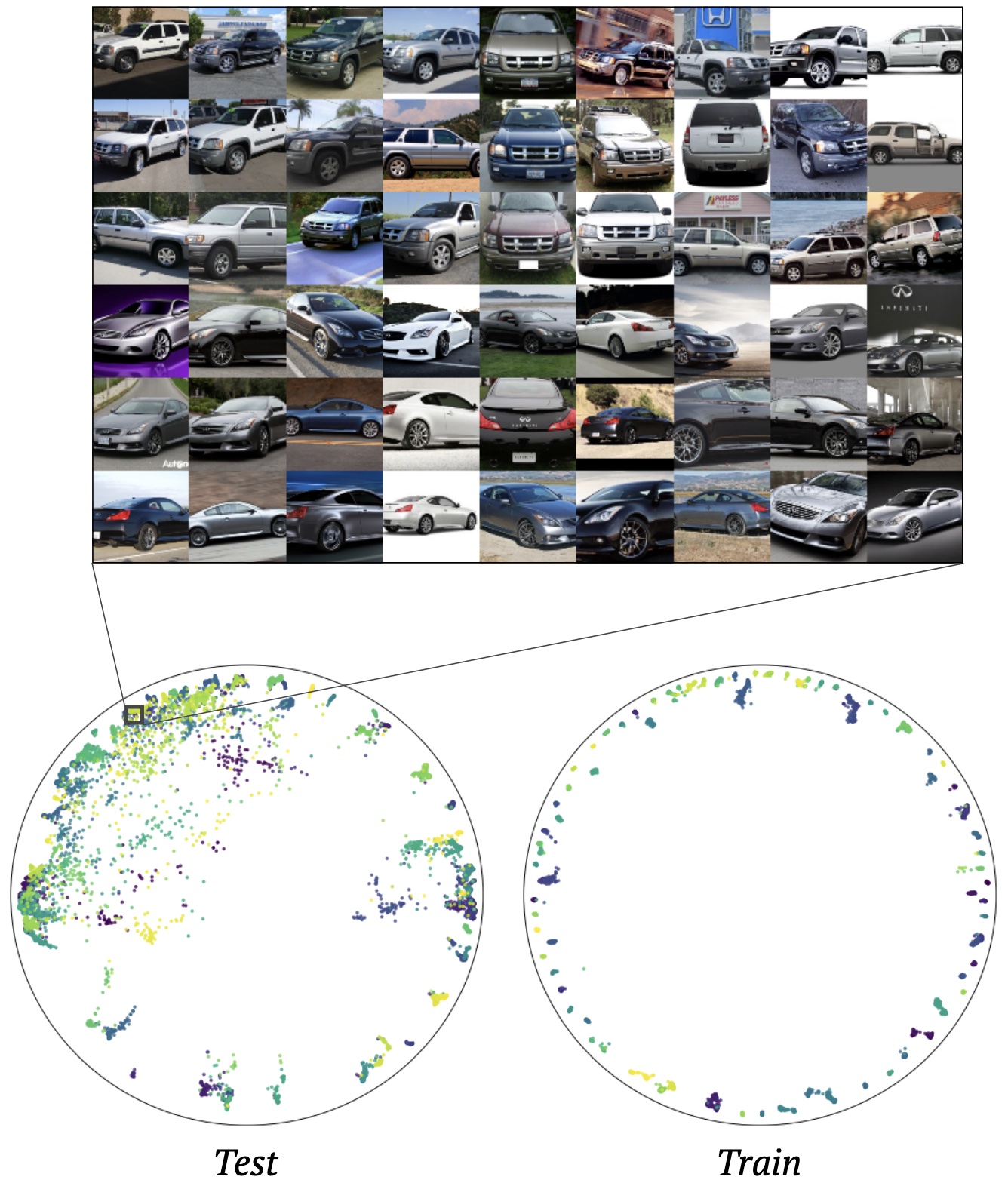}
\caption{\textbf{Embeddings of hyperbolic vision transformers} cluster samples based on their label towards the boundary of the Poincar\'e ball, while simultaneously exhibiting latent hierarchical relations. Image courtesy of \cite{ermolov2022hyperbolic}.}
\label{fig:supervised-ermolov}
\end{figure}

\subsection{Sample-to-sample learning}
Lastly, a number of recent works have investigated hyperbolic learning by contrasting between samples.

\paragraph{Hyperbolic metric learning.}
\cite{ermolov2022hyperbolic} investigate the potential of hyperbolic embedding for metric learning. In metric learning, the de facto solution is to match representations of sample pairs based on embeddings given by a pre-trained encoder. Rather than relying on Euclidean distances and contrastive learning for optimization, they propose a hyperbolic pairwise cross-entropy loss. Given a dataset with $|\mathcal{Y}|$ classes, each batch samples two samples from each category, \ie $K = 2 \cdot |\mathcal{Y}|$. Then the loss function for a positive pair with the same class label is given as:
\begin{equation}
\ell_{ij} = - \log \frac{\exp(-D(g(x_i), g(x_j)) / \tau)}{\sum_{k=1}^{K} \exp(-D(g(x_i), g(x_k)) / \tau)},
\end{equation}
where $D(\cdot, \cdot)$ can be either a hyperbolic or a cosine distance and $\tau$ denotes a temperature hyperparameter. This loss is computed over all positive pairs $(i,j)$ and $(j,i)$ in a batch. Using supervised \citep{dosovitskiy2020image} and self-supervised \citep{caron2021emerging} vision transformers as encoders, hyperbolic metric learning consistently outperforms Euclidean alternatives and sets state-of-the-art on fine-grained datasets. Figure \ref{fig:supervised-ermolov} shows a 2D projection of the embeddings learned with hyperbolic metric learning on vision transformers, where classes are grouped towards the boundary and latent hierarchical neighborhood relations emerge. 

\newpaper{Hyperbolic metric learning has shown to be effective to overcome overfitting and catastrophic forgetting in few-shot class-incremental learning tasks, explored by ~\cite{cui2022rethinking}. This is done by adding a metric learning loss as a part of the distillation in continual learning. They also propose a hyperbolic version of Reciprocal Point Learning~\citep{chen2020learning} to provide extra-class space for known categories in the few-shot learning stage.}\newpapertwo{~\cite{yan2023adaptive} also explore hyperbolic metric learning, incorporating noise-insensitive and adaptive hierarchical similarity to handle noisy labels and multi-level relations. \cite{kim2022hier} add a hierarchical regularization term on top of the metric learning approaches, with the goal of learning hierarchical ancestors in hyperbolic space without any annotation. Hyperbolic metric learning is furthermore effective in semantic hashing~\citep{amin2022deep}, face recognition via large-margin nearest-neighbor learning \citep{trpin2022face}, and multi-modal alignment given videos and knowledge graph~\citep{guo2021multi}.}

\lastpaper{Following the progress of large language models and the success of vision-language models (\eg CLIP~\citep{radford2021learning}) in multimodal representation learning, \cite{desai2023hyperbolic} propose a hyperbolic image-text representation. The proposed method first processes the input image and text using two separate encoders. Then, the generated embedding is projected into the hyperbolic space, and training is performed using a contrastive and entailment loss. The paper shows that the proposed approach outperforms the Euclidean CLIP as it is capable of capturing hierarchical multimodal relations in hyperbolic space.}

\paragraph{Hyperbolic set-based learning.}
Where sample-to-prototype and sample-to-sample approaches compare samples to individual elements, some works have shown that set-based and group-based distances are more effective and robust. \cite{ma2022adaptive} introduce an adaptive sample-to-set distance function in the context of few-shot learning. Rather than aggregating support samples to a single prototype, an adaptive sample-to-set approach is proposed to increase the robustness to the outliers. The sample-to-set function is a weighted average of the distance from the query to all support samples, where the distance is calculated with a small network over the feature maps of the query and support samples. This approach benefits few-shot learning, especially when dealing with outliers.

In the context of metric learning, \cite{zhang2021learning} argue that sample-to-sample learning is computationally expensive, while sample-to-prototype learning is less accurate. They propose a hybrid strategy based on grouplets. Each grouplet is a random subset of samples and the set of grouplets is matched with prototypes through a differentiable optimal transport. Akin to \cite{ermolov2022hyperbolic}, they show that using hyperbolic embedding spaces improved metric learning on fine-grained datasets. Moreover, they provide empirical evidence that other metric-based losses benefit from hyperbolic embeddings, highlighting the general utility of hyperbolic space for metric learning.
\begin{figure*}[t]
\centering
\includegraphics[width=\textwidth]{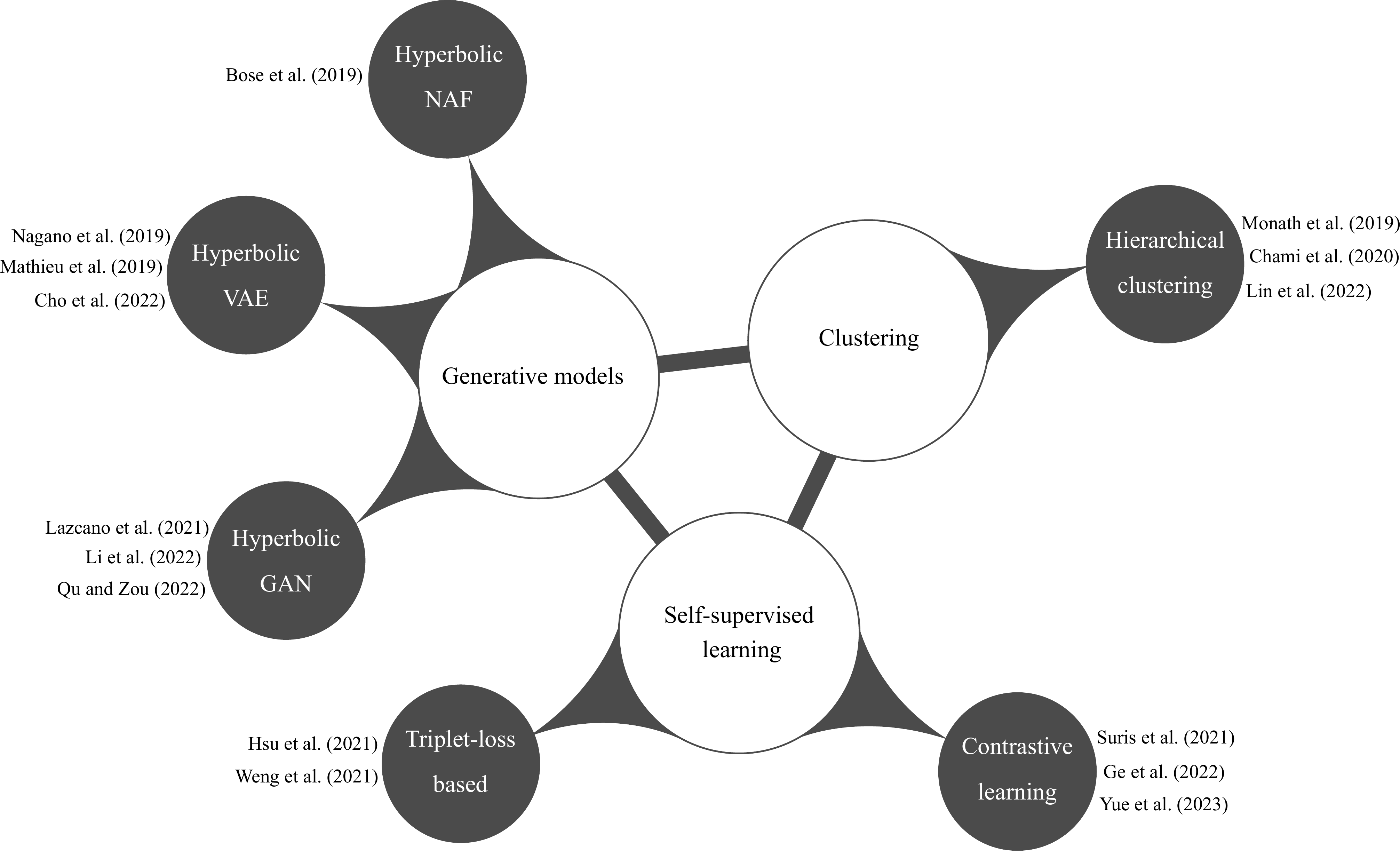}
\caption{\textbf{The three major methods for unsupervised hyperbolic learning in computer vision.} Current literature performs unsupervised learning in hyperbolic space using  (i) generative models, (ii) clustering, (iii) self-supervised learning. }
\label{fig:unsupervised-venn}
\end{figure*}.

\section{Unsupervised hyperbolic visual learning}
\label{sec:unsupervised}
Hyperbolic learning has been actively researched in the unsupervised domain of computer vision. We identify three dominant research directions in which hyperbolic deep learning has found success: generative learning, clustering, and self-supervised learning. Below, each is discussed separately.

\subsection{Generative approaches}

\subsubsection{Hyperbolic VAEs}

Variational autoencoders (VAEs) \citep{kingma2013auto, rezende2014stochastic} with hyperbolic latent space have been used to learn representations of images. \citet{nagano2019wrapped} propose the hyperbolic wrapped normal distribution and derive algorithms for both reparametrizable sampling and computing the probability density function. They then derive a hyperbolic $\beta$-VAE \citep{higgins2017beta} using the wrapped normal function as the prior and posterior, replacing the usual (Euclidean) Gaussian distribution. The wrapped normal distribution in a manifold $\mathcal{M}$ is the pushforward measure under the exponential map $\exp_\mathcal{M}$. Thus, a sample $z$ can be obtained as \citep{mathieu2019continuous}:
\begin{align}
    z = \exp_\mu^\mathcal{M}\left( G(\mu)^{-1/2} v\right), v \sim \mathcal{N}(\cdot | 0, \Sigma)
\end{align}
where $\exp_\mu^\mathcal{M}$ is the exponential map of $\mathcal{M}$ at $\mu$ and $G$ is the matrix representation of the metric of $\mathcal{M}$. To accommodate the geometry of the latent space, exponential and logarithmic maps were added at the end of the VAE encoder and before the start of the VAE decoder, respectively. In order to train their hyperbolic VAE with the typical evidence lower bound, \citet{nagano2019wrapped} compute the density of the wrapped normal distribution using the change-of-variables formula. Since their sampling algorithm required the exponential and parallel transport maps, \citet{nagano2019wrapped} compute the log-determinants and inverses of these maps in order to apply the change-of-variables formula. \citet{nagano2019wrapped} then use their VAE to learn representations of MNIST and Atari 2600 Breakout screens. On MNIST, Hyperbolic representations outperform Euclidean representations at low latent dimensions but were overtaken starting at dimension 10. 

\citet{mathieu2019continuous} extend the work of \citet{nagano2019wrapped} by introducing the Riemannian normal distribution and deriving reparametrizable sampling schemes for both the Riemannian normal and wrapped normal using hyperbolic polar coordinates. The Riemannian normal views the Euclidean normal distribution as the distribution minimizing the entropy for a given mean and standard deviation and defines a new normal distribution on hyperbolic space with this property:
\begin{align}
    \mathcal{N}_{\mathcal{M}}^R(z | \mu, \sigma^2) = \frac{1}{Z^R} \exp \left(-\frac{d_\mathcal{M}(\mu, z)^2}{2\sigma^2} \right)
\end{align}
where $Z^R$ is a normalizing constant.  
\citet{mathieu2019continuous} additionally introduce the use of a gyroplane layer as the first layer of the decoder, following \citet{ganea2018hyperbolic}. Noting that an Euclidean affine transform can be written as 
\[f_{a, p}(z) = \text{sign}(\langle a, z-p\rangle)||a||d_E(z, H_{a, p})\]
where $H_{a, p} = \{z \in \mathbb{R}^n | \langle a, z-p \rangle = 0\}$ is the decision hyperplane, they replace each piece of the formula with its hyperbolic counterpart to obtain
\begin{align}
\label{eqn:gyroplane}
    f_{a, p}^c(z) = \text{sign}(\langle a, \log_p^c(z) \rangle_p)||a||_p d_p^c(z, H_{a, p}^c)
\end{align}
where all $H_{a, p}^c = \{z \in \mathbb{H} | \langle a, \log_p^c(z) \rangle = 0\}$. The closed-form formula for the distance term in the Poincar\'{e} ball is
\begin{align}
\label{eqn:hyp-distance-to-hyperplane}
    d_p^c(z, H_{a, p}^c) = \frac{1}{\sqrt{c}} \sinh^{-1}\left(\frac{2\sqrt{c}|\langle -p \oplus_c z, a \rangle|}{(1 - c||-p \oplus_c z ||^2)||a||} \right)
\end{align}
\citet{mathieu2019continuous} also use their hyperbolic VAE to learn representations of MNIST and find that using both the Riemannian normal and the gyroplane layer improve test log-likelihoods, especially at low latent dimensions. 

\begin{figure}[t]
\centering
\includegraphics[width=\linewidth]{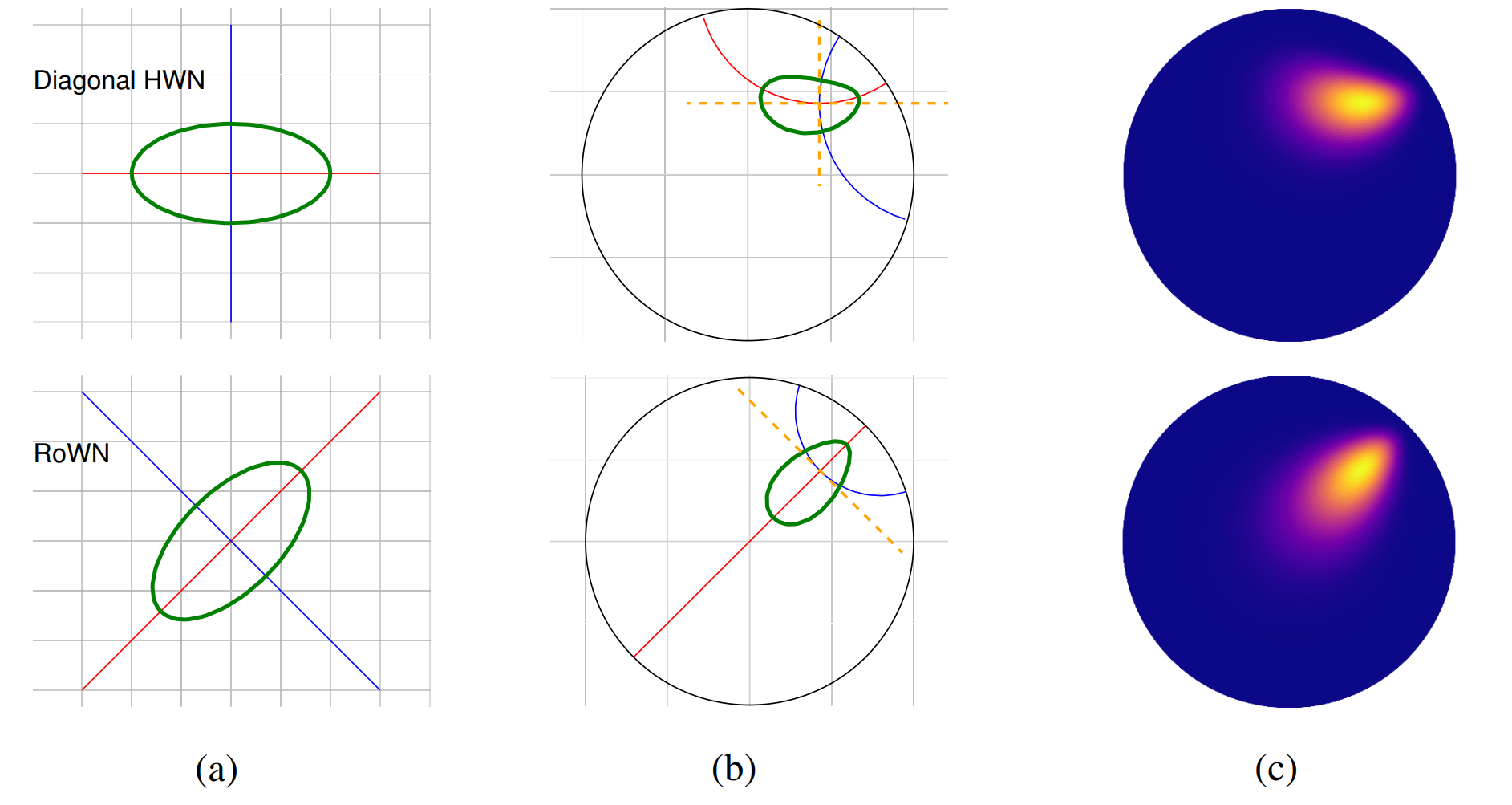}
\caption{The \textbf{standard hyperbolic wrapped normal} (top) and \textbf{rotated hyperbolic wrapped normal} (bottom). In (a), the principal axes of the normal distribution are illustrated. In (b), the principal axes of the transported normal distribution are visualized. The density of the two distributions are visualized in (c). Image courtesy of \cite{cho2022rotated}.}
\label{fig:rotated-wn}
\end{figure}

\cite{cho2022rotated} extend the previous two works by proposing a new version of the hyperbolic wrapped normal distribution (HWN). Their primary observation is that for the wrapped normal distribution, the principal axes of the distributions are not aligned with the local standard axes, see Figure \ref{fig:rotated-wn}. They propose a new sampling process that fixes the alignment of the principal axes, resulting in a new distribution which they call the rotated hyperbolic wrapped normal (RoWN). Given a mean $\mu$ in the Lorentz model of hyperbolic geometry and a diagonal covariance matrix $\Sigma$, samples from the RoWN distribution are sampled as follows:
\begin{enumerate}
    \item Find the rotation matrix $R$ that rotates the $x$-axis $x = \left([\pm 1, \ldots, 0] \right)$ to $y = \mu_{1:}$. We can compute $R$ as
    \begin{align}
        R = I + (y^Tx - x^Ty) + \frac{(y^Tx - x^Ty)^2}{1 + \langle x, y \rangle}
    \end{align}
    \item Rotate $\Sigma$ by $R$: $\hat{\Sigma} = R\Sigma R^T$
    \item Now sample as in the usual hyperbolic wrapped normal: sample $v \sim \mathcal{N}(\mathbf{0}, \hat{\Sigma})$ and then map it to hyperbolic space as follows: $\exp_\mu(\text{PT}_{\mathbf{0} \to \mu}([0, v]))$
\end{enumerate}
\cite{cho2022rotated} find that RoWN outperforms HWN in a variety of settings, such as the Atari 2600 Breakout image generation experiment first examined in \cite{nagano2019wrapped}.  
 
\subsubsection{Hyperbolic GANs}

Using the intuition that images are organized hierarchically, several works have proposed hyperbolic generative adversarial networks (GANs). \citet{lazcano2021hgan} propose a hyperbolic GAN which replaces some of the Euclidean layers in both the generator and discriminator with hyperbolic layers \citep{ganea2018hyperbolic2} with learnable curvature. \citet{lazcano2021hgan} propose hyperbolic variants of the original GAN \citep{goodfellow2020generative}, the Wasserstein GAN WGAN-GP \citep{gulrajani2017improved} and conditional GAN CGAN \citep{mirza2014conditional}. The paper finds that their best configurations of Euclidean and hyperbolic layers generally improved the Inception Score \citep{salimans2016improved} and Frechet Inception Distance \citep{heusel2017gans} on MNIST image generation, with the best improvements in the GAN architecture. The best learned curvatures are close to zero. Unlike other hyperbolic generative models (VAEs and normalizing flows), good results are observed at large latent dimensions.

\cite{qu2022autoencoding} propose HAEGAN, a hyperbolic autoencoder and GAN framework in the Lorentz model $\mathbb{L}$ (also known as the hyperboloid model), of hyperbolic geometry. The GAN is based on the structure of WGAN-GP \citep{arjovsky2017wasserstein, gulrajani2017improved}. The structure of HAEGAN consists of an encoder, which takes in real data and generates real representations, and a generator, which takes in noise and generates fake representations. A critic is trained to distinguish between the two representations, and a decoder takes the fake representations and produces the final generated object. \citet{qu2022autoencoding} generalize WGAN-GP to hyperbolic space using three operations: the first is the hyperbolic linear layer is $\texttt{HLinear}_{n, m}: \mathbb{L}_K^n \to \mathbb{L}_K^m$
of \cite{chen2021fully}
, the second the hyperbolic centroid distance layer $\texttt{HCDist}_{n, m}(x): \mathbb{L}_K^n \to \mathbb{R}^m$ of \cite{liu2019hyperbolic}, and the third a a new Lorentz concatenation layer:
\begin{align}
    \texttt{HCat}(\{x_i\}_{i = 1}^N) = \left[\sqrt{\sum_{i = 1}^N x_{i_t}^2 + (N - 1)/K}, x_{1_s}^\top, \ldots, x_{1_s}^\top \right]^\top
\end{align}
Compared to previous work \citet{shimizu2021hyperbolic}, the $\texttt{HCat}$ layer has the advantage of always having bounded gradients. \citep{shimizu2021hyperbolic}. Compared to \citet{lazcano2021hgan}, HAEGAN shows improved results on MNIST image generation. 

\begin{figure}[t]
\centering
\includegraphics[width=\linewidth]{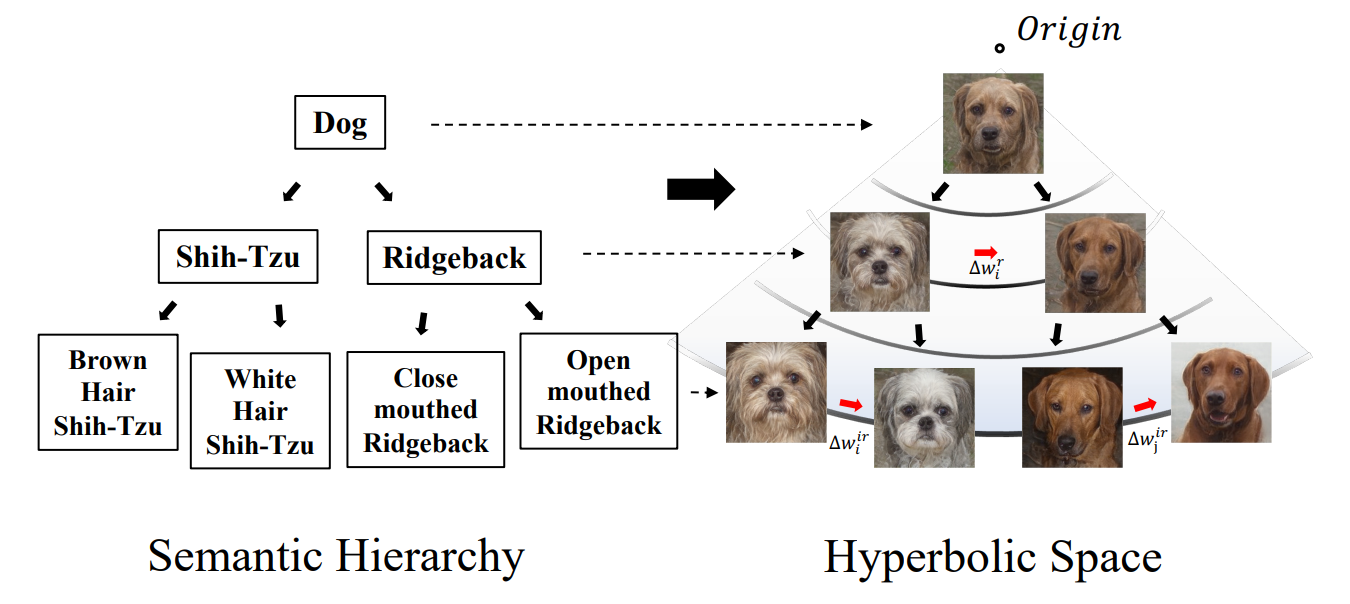}
\caption{\textbf{Hierarchical attribute editing in hyperbolic space} is possible due to hyperbolic space's ability to encode semantic hierarchical structure within image data. Changing the high-level, category-relevant details (closest to the origin) changes the category, while changing low-level (farthest from the origin), category-irrelevant attributes varies images within categories. Image courtesy of \cite{li2022euclidean}.}
\label{fig:li-few-shot-img-gen}
\end{figure}

\citet{li2022euclidean} propose a hyperbolic method for few-shot image generation. The main idea is that hyperbolic space encodes a semantic hierarchy, where the root of the hierarchy (\ie at the center of hyperbolic space) is a category, \eg dog. At lower levels, we have more fine-grained separations, such as subcategories, \eg Shih-Tzu and Ridgeback dogs. Finally, at the lowest level, there are category-irrelevant features, \eg the hair color or pose of the dog (see Figure \ref{fig:li-few-shot-img-gen}). This method builds on the Euclidean pSp method \citep{richardson2021encoding} for image-to-image translation. The pSp method uses a feature pyramid to extract feature maps and uses a set of projection heads on these feature maps to produce each of the style vectors required by StyleGAN \citep{karras2019style, karras2020analyzing}, which is commonly denoted the $\mathcal{W}^+$-space. Image-to-image translation can then be done by editing or replacing style vectors. \cite{li2022euclidean} generalize to hyperbolic space by mapping the output of a frozen, pre-trained pSp encoder to hyperbolic space and then back to the $\mathcal{W}^+$-space of style vectors, and then feeding the style vectors into a frozen, pre-trained StyleGAN. Projection to hyperbolic space is done using the Mobius layer $f^{\otimes c}$ of \cite{ganea2018hyperbolic}, with the full projection layer having the form
\begin{align}
    z_{\mathbb{D}i} = f^{\otimes c} ( \exp_0^c( \texttt{MLP}_E(\mathbf{w}_i)))
\end{align}
with mapping back to the $\mathcal{W}^+$-space achieved by a logarithmic map plus an MLP. \cite{li2022euclidean} supervise the hyperbolic latent space with a hyperbolic classification loss based on the multinomial logistic regression formulation of \cite{ganea2018hyperbolic}. After calculating the probabilities, the loss function is just negative log-likelihood as
\begin{align}
    \mathcal{L}_{\mathrm{hyper}} = -\frac{1}{N} \sum_{i = 1}^N \log(p_n)
\end{align}
The full loss function is the pSp loss function plus this term, excluding a specific facial reconstruction loss used by the pSp method, since \cite{li2022euclidean} do not focus on face generation. \cite{li2022euclidean} perform image generation as follows: given an image $x_i$, the image is embedded in hyperbolic space with representation $g_{\mathbb{D}}(x_i)$ and is rescaled to the desired radius (\ie fine-grained-ness) $r$. A random vector is then sampled from the seen categories and a point is taken on the geodesic between the two points. \citet{li2022euclidean} find that their method is competitive with state-of-the-art methods and show promise for image-to-image transfer.  

\subsubsection{Hyperbolic Normalizing Flows}

\begin{figure}[t]
\centering
\includegraphics[width=\linewidth]{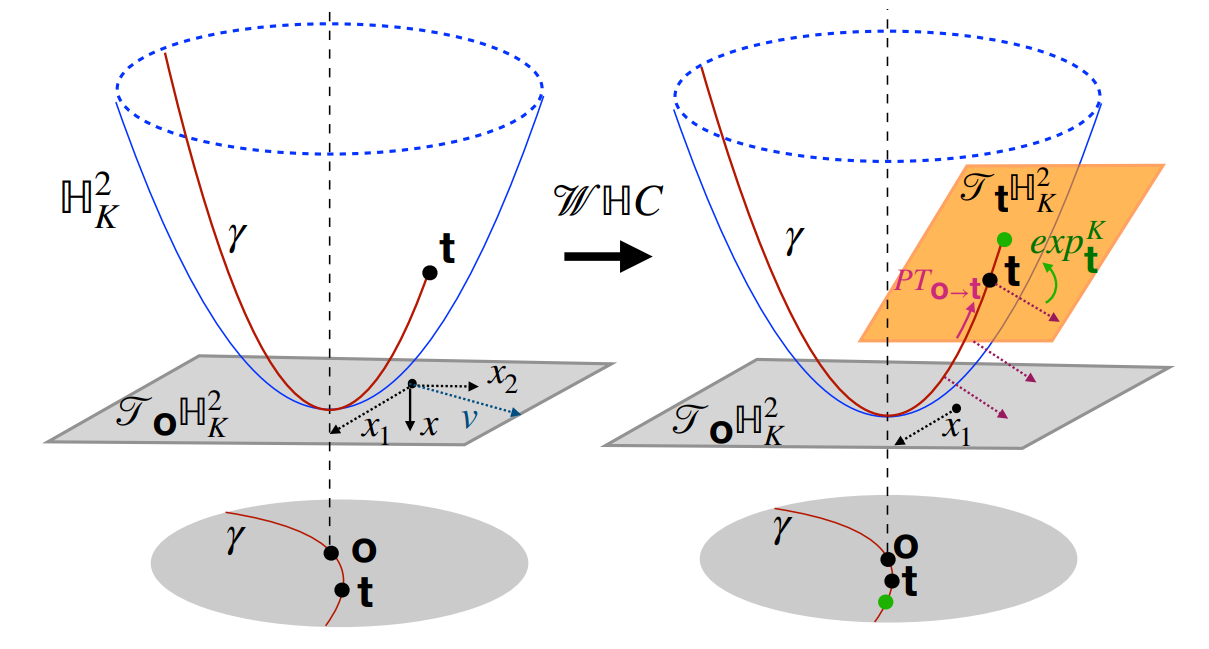}
\caption{The left figure shows the partitioning step of wrapped hyperbolic coupling, and the right figure shows how the vector is transformed, transported, and projected back to hyperbolic space. Image courtesy of \cite{bose2020latent}. }
\label{fig:bose-hyp-naf}
\end{figure}

\citet{bose2020latent} propose a hyperbolic normalizing flow that generalizes the Euclidean normalizing flow RealNVP \citep{dinh2016density} to hyperbolic space. They propose two types of hyperbolic normalizing flows: the first, which they call tangent coupling, which carries out the coupling layer of RealNVP in the tangent space at the hyperbolic origin $o$:
\begin{align}
    &\Tilde{f}^{\mathcal{T}C}(\Tilde{x}) = \begin{cases}
    \Tilde{z}_1 = \Tilde{x}_1 \\
    \Tilde{z}_2 = \Tilde{x}_2 \odot \sigma(s(\Tilde{x}_1)) + t(\Tilde{x}_1)
    \end{cases} \\
    &f^{\mathcal{T}C}(x) = \exp_o^K(\Tilde{f}^{\mathcal{T}C}(\log_o^K(x)))
\end{align}
where $s, t$ are neural networks and $\sigma$ is a pointwise non-linearity.

The wrapped hyperboloid extends tangent coupling by using parallel transport to map intermediate vectors from the tangent space of the origin to the tangent space of another point in hyperbolic space (see Figure \ref{fig:bose-hyp-naf}):
\begin{align}
    \Tilde{f}^{\mathcal{W}\mathbb{H}C}(\Tilde{x}) &= \begin{cases}
    \Tilde{z}_1 = \Tilde{x}_1 \\
    \Tilde{z}_2  = \log_o^K \left( \exp_{t(\Tilde{x}_1)}^K \left(\mathrm{PT}_{o \to t(\Tilde{x}_1)}(v) \right) \right)
    \end{cases} \\
    v &= \Tilde{x}_2 \odot \sigma(s(\Tilde{x }_1)) \\
    f^{\mathcal{W}\mathbb{H}C}(x) &= \exp_o^K(\Tilde{f}^{\mathcal{W}\mathbb{H}C}(\log_o^K(x)))
\end{align}
Compared to tangent coupling, wrapped hyperbolic coupling allows the flow to leverage different parts of the manifold instead of just the origin. The paper also derives the inverse and Jacobian determinants of the two flows. As is the case for hyperbolic VAEs, \citet{bose2020latent} also benchmark on MNIST, and find a similar trend as \citet{nagano2019wrapped}: the performance of hyperbolic models exceed that of the equivalent Euclidean model at low dimension, but as early as latent dimension 6 Euclidean models overtake hyperbolic models in performance. \citet{bose2020latent} find that hyperbolic normalizing flows outperform hyperbolic VAEs at these low latent dimensions. 

\subsection{Clustering}

Due to the close relationship between hyperbolic space, hierarchies, and trees, several works have explored hierarchical clustering using hyperbolic space. \citet{monath2019gradient} propose to perform hierarchical clustering using hyperbolic representations. Given a dataset $\mathcal{D} = \{x_i\}_{i = 1}^N$, \cite{monath2019gradient} require a hyperbolic representation at the edge of the Poincar\'{e} disk $\mathbb{D}^d$ for each data point $x_i \in \mathcal{D}$, which becomes the leaves of the hierarchical clustering. The method of \cite{monath2019gradient} creates a hierarchical clustering by optimizing the hyperbolic representations for a fixed number of internal nodes. Parent-children dissimilarity between a child representation $z_c$ and a parent representation $z_p$ is measured by 
\begin{align}
    d_{cp}(z_c, z_p) = d_\mathbb{D}(z_c, z_p)(1 + \max\{||z_p||_\mathbb{D} - ||z_c||_\mathbb{D}, 0\})
\end{align}
which encourages children to have larger norms than their parents. A discrete tree can then be extracted as follows:
\begin{align}
    \texttt{Parent}(z_c) = \argmin_{||z_p|| < ||z_c||} d_{cp}(z_c, z_p)
\end{align}
The internal node observations are supervised by two losses: first, a hierarchical clustering loss based on Dasgupta's cost \citep{dasgupta2016cost} and a continuous extension due to \citet{wang2018improved} that reformulates the loss in terms of last common ancestors (LCAs), and second, a parent-child margin objective that encourages parent nodes to have smaller norm than their children. 

Suppose $\mathcal{D}$ has pairwise similarities $\{w_{ij}\}_{i, j \in [N]}$. A hierarchical clustering of $\mathcal{D}$ is a rooted tree $T$ such that each leaf is a data point. For leaves $i, j \in T$, denote their LCA by $i \vee j$, the subtree rooted at $i \vee j$ by $T[i \vee j]$, and the leaves of $T[i \vee j]$ by $\texttt{leaves}(T[i \vee j])$. Finally, let relation $\{i, j|k\}$ holds if $i \vee j$ is a descendant of $i \vee j \vee k$. Then Dasgupta's cost can be formulated as 
\begin{align}
\label{eqn:dasguptas-cost}
    C_{\mathrm{Dasgupta}}(T; w) = \sum_{ij} w_{ij} |\texttt{leaves}(T[i \vee j])|
\end{align}
\cite{wang2018improved} show that 
\begin{align}
\label{eqn:wangwang}
    C_{\mathrm{Dasgupta}}(T; w) &= \sum_{ijk} [w_{ij} + w_{ik} + w_{jk} - w_{ijk}(T; w)] \\
    & + 2 \sum_{ij} w_{ij}
\end{align}
where 
\begin{align}
    w_{ijk}(T; w) = w_{ij}\mathbbm{1}[\{i, j|k\}] + w_{ik}\mathbbm{1}[\{i, k|j\}] + w_{jk}\mathbbm{1}[\{j, k|i\}]
\end{align}

The margin parent-child dissimilarity is given as 
\begin{align}
    d_{cp}(z_c, z_p; \gamma) = d_\mathbb{D}(z_c, z_p)(1 + \max\{||z_p||_\mathbb{D} - ||z_c||_\mathbb{D} + \gamma, 0\})
\end{align}
and the total margin objective is 
\begin{align}
    \mathcal{L}_{cp} = \sum_{z_c} d_{cp}(z_c, \texttt{Parent}(z_c); \gamma)
\end{align}
The embedding is alternately optimized between the clustering objective and the parent-child objective. Optimization of the hyperbolic parameters is done via the method of \cite{nickel2017poincare}. Using this method, \cite{monath2019gradient} are able to embed ImageNet using representations taken from the last layer of a pre-trained Inception neural network. 

Similar to \cite{monath2019gradient}, \cite{chami2020trees} base their method on Dasgupta's cost (Equation \ref{eqn:dasguptas-cost}) and Wang and Wang's (Equation \ref{eqn:wangwang}) reformulation in terms of LCAs. \cite{chami2020trees} define the LCA of two points in hyperbolic space to be the point on the geodesic connecting the two points that are closest to the hyperbolic origin, and provide a formula to calculate this point in the Poincar\'{e} disk $\mathbb{D}$. This formula allows Equation \ref{eqn:wangwang} to be directly optimized by replacing the $w_{ijk}(T; w)$ terms with its continuous counterpart. A hierarchical clustering tree can then be produced by iteratively merging the most similar pairs, where similarity is measured by their hyperbolic LCA distance from the origin. Unlike the method of \cite{monath2019gradient}, \cite{chami2020trees} do not require hyperbolic embeddings to be available, and optimize the hyperbolic embeddings of the whole tree, not just the leaves. 

\cite{lin2022contrastive} propose a neural-network based framework for the hierarchical clustering of multi-view data. The framework consists of two steps: first, improving representation quality via reconstruction loss, contrastive learning between different views, and a weighted triplet loss between positive examples and mined hard negative examples, and second, applying the hyperbolic hierarchical clustering framework of \cite{chami2020trees}. 

The contrastive loss in \cite{lin2022contrastive} is the usual contrastive loss (see following section) where positive examples are views from the same object and negative examples are views from different objects. The weighted triplet loss is
\begin{align}
    \mathcal{L}_m = \frac{1}{N}\sum_{i = 1}^N w^m(a_i, p_i)[m + ||a_i - p_i ||^2_2 - ||a_i - n_i ||^2_2]_+
\end{align}
where $a_i$ refer to the anchor points, $p_i$ are the positive examples, and $n_i$ are the negative examples. Positive and negatives examples are mined based on the method of \cite{iscen2017efficient}, which measures the similarity of a pair of points based on estimating the data manifold using $k$-nearest neighbors graphs. \cite{lin2022contrastive} apply their method to perform multi-view clustering for a variety of multi-view image datasets. 

\subsection{Self-supervised learning}

In Section \ref{triplet-based}, we describe methods for hyperbolic self-supervision which are primarily based on triplet losses, and in Section \ref{contrastive-learning} we discuss methods for hyperbolic self-supervision which are primarily based on contrastive losses. 

\subsubsection{Hyperbolic self-supervision}
\label{triplet-based}

Based on the idea that biomedical images are inherently hierarchical, \citet{hsu2021capturing} propose to learn patch-level representations of 3D biomedical images using a 3D hyperbolic VAE and to perform 3D unsupervised segmentation by clustering the representations. \citet{hsu2021capturing} extend the hyperbolic VAE architecture of \citet{mathieu2019continuous} using a 3D convolutional encoder and decoder as well as gyroplane convolutional layer that generalizes the Euclidean convolution with the gyroplane layer of \citet{ganea2018hyperbolic} (See Equations \ref{eqn:gyroplane} and \ref{eqn:hyp-distance-to-hyperplane}). In order to learn good representations, the paper proposes to use a hierarchical self-supervised loss that captures the implicit hierarchical structure of 3D biomedical images. 

To capture the hierarchical structure of 3D biomedical images, \citet{hsu2021capturing} propose that given a parent patch $\mu_p$, to sample a child patch $\mu_c$ which is a subpatch of the parent patch, and a negative patch $\mu_n$ that does not overlap with the parent patch. Then the hierarchical self-supervised loss is defined as a margin triplet loss as follows:
\begin{align}
    \mathcal{L}_{\mathrm{hierarchical}} = \max(0, d_\mathbb{D}(\mu_p, \mu_c) - d_\mathbb{D}(\mu_p, \mu_n) + \gamma)
\end{align}
This encourages the representations of subpatches to be children or descendants of the representation of the main patch, and faraway patches (which likely contain different structures) to be on other branches of the learned hierarchical representation.

To perform unsupervised segmentation, the learned latent representations are extracted and clustered using a hyperbolic k-means algorithm, where the traditional Euclidean mean is replaced with the Frechet mean. For a manifold $\mathcal{M}$ with metric $d_\mathcal{M}$, the Frechet mean of a set of points $\{z_i\}_{i = 1}^k, z_i \in \mathcal{M}$ is defined as the point $\mu$ that minimizes the squared distance to all points $z_i$: 
\begin{align}
    \mu_{\mathrm{Fr}} = \argmin_{\mu \in \mathcal{M}} \frac{1}{k} \sum_{i = 1}^k d_\mathcal{M}(z_i, \mu)^2 
\end{align}
and is one way to generalize the concept of a mean to manifolds. Unfortunately, the Frechet mean on the Poincar\'{e} ball does not admit a closed-form solution, so \citet{hsu2021capturing} compute the Frechet mean with the iterative algorithm of \citet{lou2020differentiating}. The paper finds that this strategy is effective for the unsupervised segmentation of both synthetic biological data and 3D brain tumor MRI scans \citep{menze2014multimodal, bakas2017advancing, bakas2018identifying}. 

\citet{weng2021unsupervised} propose to leverage the hierarchical structure of objects within images to perform weakly-supervised long-tail instance segmentation. To capture this hierarchical structure, \citet{weng2021unsupervised} learn hyperbolic representations which are supervised with several hyperbolic self-supervised losses. Instance segmentation is done in three stages: first, mask proposals are generated using a pre-trained mask proposal network. Mask proposals consists of bounding boxes $\{\mathcal{B}_i\}_{i = 1}^k$ and masks $\{\mathcal{M}_i\}_{i = 1}^k$. Define $x_i^{\mathrm{full}}$ to be the original image cropped to bounding box $\mathcal{B}_i$, $x_i^{\mathrm{bg}}$ to be the cropped image with the object masked out using mask $1 - \mathcal{M}_i$, and $x_i^{\mathrm{fg}}$ to be the same cropped image with the background masked out using mask $\mathcal{M}_i$. We will refer to these as the full object image, object background, and object, respectively. 

Second, hyperbolic representations of $z_i^{\mathrm{bg}} = g(x_i^{\mathrm{bg}})$, and $z_i^{\mathrm{fg}} = g(x_i^{\mathrm{fg}})$ are learned by a pre-trained feature extractor and supervised by a combination of three self-supervised losses. The representations are fixed to have latent dimension 2. The first self-supervised loss encourages representation of the object to be similar to that of the full object image and farther away from the representation of the object background:
\begin{align}
    \mathcal{L}_\mathrm{mask} = \sum_{i = 1}^k \max(0, \gamma - d(z_i^\mathrm{full}, z^\mathrm{fg}) + d(z_i^{\mathrm{full}}, z_i^{\mathrm{bg}}))
\end{align}
The second loss is a triplet loss that requires the sampling of positive and negative examples. 
\begin{align}
    \mathcal{L}_\mathrm{object} = \sum_{i = 1}^k \max(0, \gamma - d(z_i^\mathrm{fg}, \hat{z}^\mathrm{fg}) + d(z_i^{\mathrm{fg}}, \overline{z}_i^{\mathrm{fg}}))
\end{align}
The third loss is similar to the hierarchical triplet loss of \citet{hsu2021capturing} described above, except with the origin taking the place of negative samples:
\begin{align}
    \mathcal{L}_\mathrm{hierarchical} = \sum_{i = 1}^k \max(0, \gamma - d(z_i^{\mathrm{child}}, o) - d(z_i^{\mathrm{fg}}, o))
\end{align}

Finally, the representations are clustered using hyperbolic k-means clustering. Unlike \citet{hsu2021capturing}, to compute the mean they map the representations from the Poincare disk to the hyperboloid model $\mathcal{L}$ and compute the (weighted) hyperboloid midpoint proposed by \citet{law2019lorentzian}:
\begin{align}
    \mu = \frac{\sum_{i = 1}^k \nu_i x_i}{\left|||\sum_{i = 1}^k \nu_i x_i ||_\mathcal{L}\right|}
\end{align}
Compared to the Frechet mean, this mean has the advantage of having a closed-form formula, making it more computationally efficient. \cite{weng2021unsupervised} find that their method improves other partially-supervised methods on the LVIS long-tail segmentation dataset \citep{gupta2019lvis}.

\subsubsection{Hyperbolic contrastive learning}
\label{contrastive-learning}

\begin{figure}[t]
\centering
\includegraphics[width=\linewidth]{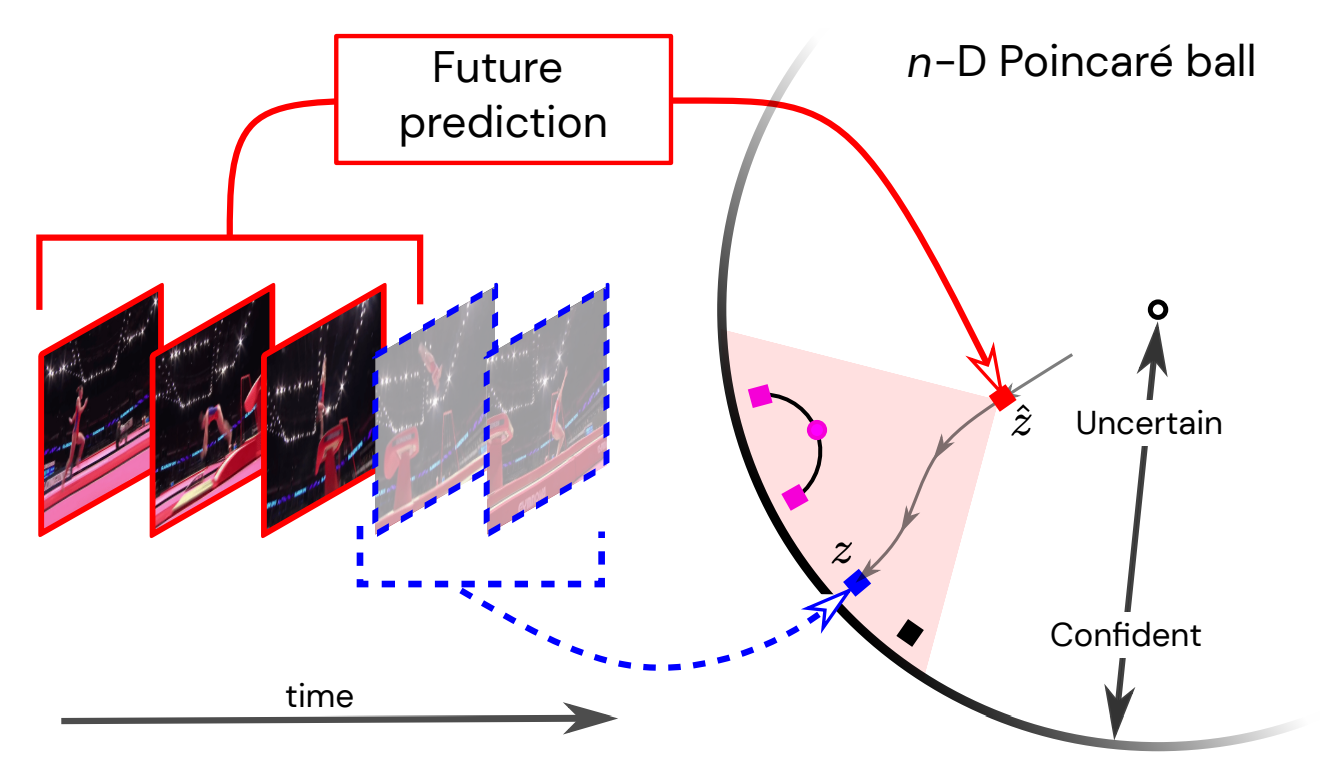}
\caption{\cite{suris2021learning} model uncertainty with hyperbolic representations. If the model is uncertain, it can predict an abstraction of all possible actions (red square), and if it is certain it can predict a more specific action (blue square). The pink circle shows how computing the mean of two representations (pink squares) increases the generality. Image courtesy of \cite{suris2021learning}.}
\label{fig:suris-pull-fig}
\end{figure}

Hyperbolic contrastive learning methods have also been proposed. \citet{suris2021learning} propose to learn hyperbolic representations for video action prediction because of their ability to combine representing hierarchy and giving a measure of uncertainty (See Figure \ref{fig:suris-pull-fig}). \cite{suris2021learning} learn an action hierarchy where more abstract actions are near the origin of the Poincar\'{e} disk and more fine-grained actions are near the edge. If the preceding video frames are ambiguous, this hierarchical representation allows the ability to predict a more general parent category of action (\eg greeting) instead of having to predict more fine-grained child categories of action (\eg handshake or high-five). The parent of two actions is computed as the hyperbolic mean of their hyperbolic representations, which \citet{suris2021learning} compute as the midpoint of the geodesic connecting the two representations. \cite{suris2021learning} propose a two-stage framework for video action prediction which consists first of contrastive pre-training hyperbolic representations, then freezing the representations and training a linear classifier for action prediction.  

Self-supervised pre-training proceeds as follows: let $x_t$ be a frame of the video, and a representation $z_t = f(x_t)$ is produced by an encoder $f$. The pretext task is to predict the representation $z_{t + \delta}$ of a clip $\delta$ frames into the future. The model produces an estimate $\hat{z}_{t + \delta} = \phi(c_t, \delta)$, where $c_t = g(z_1, \ldots, z_t)$ is an encoding of all past video frames. All function $f, g, \phi$ are parameterized by a neural network. The training is supervised by a contrastive loss:
\begin{align}
    \mathcal{L} = -\sum_i \left[\log \frac{\exp(-d_\mathbb{D}^2(\hat{z}_i, z_i))}{\sum_j \exp(-d_\mathbb{D}^2(\hat{z}_i, z_j))} \right]
\end{align}
which encourages the positive pairs $\hat{z}_i, z_i$ to have similar representations while pushing $\hat{z}_i$ from the representations of all negative examples $z_j$. One key feature of this loss is that under the presence of uncertainty, say when actions $a, b$ are probable, $\mathcal{L}$ is minimized by predicting the midpoint on the geodesic connecting $a, b$, which is equivalent to moving one level up the hierarchy to the parent of $a, b$. 

\begin{figure}[t]
\centering
\includegraphics[width=\linewidth]{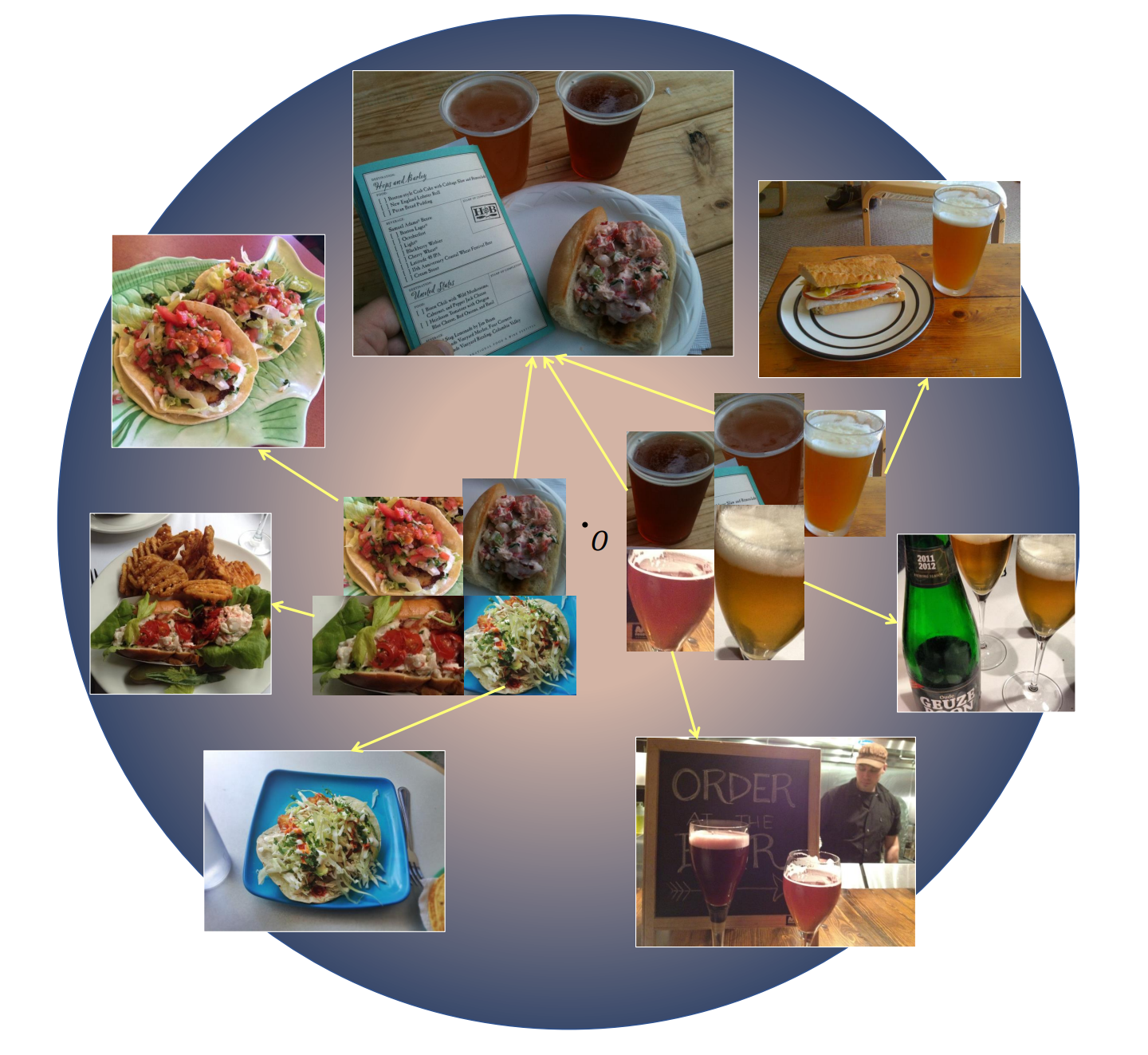}
\caption{The learned hierarchy of \cite{ge2022hyperbolic} has objects near the origin of the Poincar\'{e} disk and scenes near the edge of hyperbolic space. Image courtesy of \cite{ge2022hyperbolic}. }
\label{fig:ge-hcl}
\end{figure}

\cite{ge2022hyperbolic} propose to improve contrastive learning by incorporating the hierarchical structure of images with a scene-object hierarchy (see Figure \ref{fig:ge-hcl}). \cite{ge2022hyperbolic} use a hyperbolic version of the MoCo architecture \citep{he2020momentum}, which the authors call HCL.  \citet{ge2022hyperbolic} extend the MoCo architectures in several ways: first, unlike previous works for visual contrastive learning, HCL requires that object regions be extracted from the input image. Secondly, a hyperbolic backbone along with a corresponding momentum encoder is added to MoCo's Euclidean backbone and its momentum encoder. The Euclidean backbone and momentum encoder are trained the same way as in \cite{he2020momentum}, but the inputs are not images but the extracted object regions. The hyperbolic branch takes as input a scene region $u$ and an object region $v$ that is a subregion of the scene $u$, and negative objects $\mathcal{N}_u = \{n_1, \ldots, n_k\}$ that are not subregions of the scene $u$. Let the representations of $u, v, n_j$ be $z_u, z_v, z_j$, respectively. The hyperbolic branch is then trained with a contrastive loss with hyperbolic distance as the similarity measure: 
\begin{align}
    \mathcal{L}_{\mathrm{hyp}} = -\log \frac{\exp\left(-d_\mathbb{D}(z_u, z_v)/\tau \right)}{\exp\left(-d_\mathbb{D}(z_u, z_v)/\tau \right) + \sum_j\exp\left(-d_\mathbb{D}(z_u, z_j)/\tau \right)}
\end{align}
where $\tau$ is a temperature parameter. This loss encourages representations to form a scene-object hierarchy where scenes have the highest norm (i.e., are at the edge of the Poincar\'{e} ball $\mathbb{D}$) and objects have the smallest norm (i.e., are at the center of $\mathbb{D}$). The paper finds that their method achieves small gains over the original MoCo and MoCo augmented with bounding box information. They also examine the representations of out-of-context objects using their method, and find that they generally have higher distance to the scene images. 

\cite{yue2023hyperbolic} propose a different method for hyperbolic contrastive learning that is based on SimCLR \citep{chen2020simple}. Like \cite{ge2022hyperbolic}, \cite{yue2023hyperbolic} replace the dot-product similarity of the contrastive loss with the hyperbolic distance:
\begin{align}
    \mathcal{L}_{hyp}^{self} =  -\sum_{i \in I} \log \frac{\exp(-d_\mathbb{D}(z_i, z_{j(i)})/\tau)}{\sum_{a \in A(i)} \exp(-d_\mathbb{D}(z_i, z_a)/\tau)}
\end{align}
but unlike \cite{ge2022hyperbolic}, they only have a hyperbolic branch and do not retain an Euclidean branch. \cite{yue2023hyperbolic} also propose to extend the supervised contrastive learning method SupCon \citep{khosla2020supervised} in the same way. \cite{yue2023hyperbolic} also propose to train an adversarially robust contrastive learner that extends the Robust Contrastive Learning (RoCL) \citep{kim2020adversarial} method to hyperbolic space by replacing the Euclidean contrastive losses in RoCL's adversarial training loss with their hyperbolic contrastive loss:
\begin{align}
    \mathcal{L}_{hyp}^{self}(\Tilde{x}, \{\Tilde{x}^+, \Tilde{x}^{adv}, \{\Tilde{x}^-\}\}) + \lambda \mathcal{L}_{hyp}^{self}(\Tilde{x}^{adv}, \Tilde{x}^+, \{\Tilde{x}^-\})
\end{align}
where $\Tilde{x}$ is a given image, $\Tilde{x}^+$ is a positive example, $\Tilde{x}^-$ is a negative example, and $\Tilde{x}^{adv}$ is an adversarial example that is within $\delta$ of $\Tilde{x}$. As in \cite{ge2022hyperbolic}, \cite{yan2021unsupervised} find that hyperbolic contrastive learning generally achieves small gains over its Euclidean counterparts. 
\section{Conclusions and future outlook}
\label{sec:conclusions}
This survey provides an overview of the current state of affairs in hyperbolic deep learning for computer vision. Based on the organization of supervised and unsupervised literature, we conclude the survey by discussing which types of problems currently benefit most from hyperbolic learning and discussing open problems for future research.

\subsection{When is hyperbolic learning most effective?}
From current works, we identify four main axes of improvement that have come with the recent shift towards learning in hyperbolic space for computer vision:
\begin{itemize}
\item[$\bullet$] \textbf{Hierarchical learning.} The inherent links between hierarchical data and hyperbolic embeddings are well known. It is therefore not all too surprising to see that a wide range of works have used hyperbolic learning to improve hierarchical objectives in computer vision. The ability to incorporate hierarchical knowledge, for example through hyperbolic embeddings or hierarchical hyperbolic logistic regression, has been utilized for several problems. Hierarchical learning in hyperbolic space can among others reduce error severity, resulting in smaller mistakes and more consistent retrieval. This is a key property for example in medical domains, where large mistakes need to be avoided at all costs.

Hierarchical learning has also shown to enable zero-shot generalization. By embedding class hierarchies in hyperbolic space and mapping examples of seen classes to their corresponding embedding, it becomes possible to generalize to examples of unseen classes. In general, hierarchical information between classes helps to structure the semantics of the task at hand, and embedding such knowledge in hyperbolic space is preferred over Euclidean space.
\item[$\bullet$] \textbf{Few-sample learning.} Few-shot learning is popular in hyperbolic deep learning for computer vision. Many works have shown that consistent improvements can be made by performing this task with hyperbolic embeddings and prototypes, both with and without hierarchical knowledge. In few-shot learning, samples are scarce when it comes to generalization, and working in hyperbolic space consistently improves accuracy. These results indicate that hyperbolic space can generalize from fewer examples, with potential in domains where examples are scarce. This is already visible in the unsupervised domain, where generative learning is better in hyperbolic space when working with constrained data sources.
\item[$\bullet$] \textbf{Robust learning.} Across several axes, hyperbolic learning has shown to be more robust. For example, hyperbolic embeddings improve out-of-distribution detection, provide a natural way to quantify uncertainty about samples, pinpoint unsupervised out-of-context samples, and can improve robustness to adversarial attacks. Robustness and uncertainty are key challenges in deep learning in general, hyperbolic deep learning can provide a natural solution to robustify networks.
\item[$\bullet$] \textbf{Low-dimensional learning.} For a lot of applications, networks, and embedding spaces need to be constrained, for example when learning on embedded devices or when visualizing data. In the unsupervised domain, hyperbolic learning consistently improves over Euclidean learning when working with smaller embedding spaces. Similarly, the embedding space in supervised problems can be substantially reduced while maintaining downstream performance in hyperbolic space. As such, hyperbolic learning has the potential to enable learning in compressed and embedded domains.
\end{itemize}

\subsection{Open research questions}
Hyperbolic learning has made an impact on computer vision with many promising avenues ahead. The field is however still in the early stages with many challenges and opportunities ahead. Three directions stand out:
\begin{itemize}
\item[$\bullet$] \textbf{Fully hyperbolic learning.} Hyperbolic learning papers in computer vision commonly share one perspective: hyperbolic learning should be done in the embedding space. For the most part, the representation learning of earlier layers is done in Euclidean space, resulting in hybrid networks. Works from neuroscience indicate that for the earlier layers in neural networks, hyperbolic space can also play a prominent role \citep{chossat2020hyperbolic}. Recently, \cite{zhang2023hippocampal} have shown that spatial relations in the hippocampus are more hyperbolic than Euclidean.

Learning deep networks fully in hyperbolic space requires rethinking all layers, from convolutions to self-attention and normalization. At the time of writing the survey, two works have made steps in this direction. \cite{bdeir2023hyperbolic} introduce a hyperbolic convolutional network in the Lorentz model of hyperbolic space. They outline how to perform convolutions, batch normalization, and residual connections. Simultaneously, \cite{van2023poincar} introduce Poincar\'e ResNet, with convolutions, residuals, batch normalization, and better network initialization in the Poincar\'e ball model. The works provide a foundation towards fully hyperbolic learning, but many open questions remain. Which model is most suitable for fully hyperbolic learning? Or do different layers work best in different models? And how can fully hyperbolic learning scale to ImageNet and beyond? Should each stage of the network have the same curvature? And how effective can hyperbolic networks become across all possible tasks compared to Euclidean networks? A lot more research is needed to answer these questions.
\item[$\bullet$] \textbf{Computational challenges.} Performing gradient-based learning in hyperbolic space changes how networks are optimized and how parameters behave. Compared to their Euclidean counterpart however, hyperbolic networks and embeddings can be numerically more unstable, with issues at the boundary of the ball, vanishing gradients, and more. Moreover, hyperbolic operations can be more involved and computationally heavy depending on the used model, leading to less efficient networks. Such computational challenges are relevant for all domains of hyperbolic learning and a broader topic that is receiving attention.
\item[$\bullet$] \textbf{Open source community.} Modern deep learning libraries are centered around Euclidean geometry. Any new researcher in hyperbolic learning, therefore, does not have the opportunity to quickly implement networks and layers to get an intuition into its workings. Moreover, any new advances have to be either implemented from scratch or imported from code repositories of other papers. What is missing is an open-source community and a shared repository that houses advances in hyperbolic learning for computer vision. Such a community and code base is vital to get further traction and attract a wide audience, including practitioners. Whether it be part of existing libraries or as a separate library, continued development of open-source hyperbolic learning code is key for the future of the field.
\item[$\bullet$] \textbf{Large and multimodal learning.} In computer vision, and Artificial Intelligence in general, there is a strong trend towards learning at large scale and learning with multiple modalities, \eg image-text or video-audio models. It is therefore a natural desire for the field to arrive at hyperbolic foundation models. While early work has shown that large-scale and/or multimodal learning is viable with hyperbolic embeddings \citep{desai2023hyperbolic}, hyperbolic foundation models form a longer-term commitment as they require solutions to all open problems mentioned above, from stable, fully hyperbolic learning to continued open source development.
\end{itemize}

\bibliographystyle{spbasic}      
\bibliography{egbib}   

\end{document}